\RequirePackage{fix-cm}
\documentclass[twocolumn]{svjour3}          %
\smartqed  %
\usepackage{graphicx}
\usepackage{amsmath}
\usepackage{multirow}
\usepackage{pifont}
\usepackage{graphicx}
\usepackage{amsmath}
\usepackage{amssymb}
\usepackage{array}
\usepackage{multirow, nicefrac}
\usepackage{pifont}
\usepackage{natbib}

\usepackage{xcolor}
\definecolor{green}{rgb}{1,0,0}

\usepackage{xspace}
\usepackage[colorlinks,linkcolor=red,anchorcolor=blue,citecolor=blue,CJKbookmarks=True]{hyperref}
\def\OurMethod{HiNAS\xspace}

\begin{document}

\title{Memory-Efficient Hierarchical Neural Architecture Search for Image Restoration\thanks{Part of this work was done when H. Zhang, H. Chen and C. Shen 
were with 
The University of Adelaide. This work was in part supported by National Natural Science Foundation of China (61871460,  61941112, 61761042), Shaanxi Provincial Key RD Program (2020KW-003), Natural Science Foundation of China of Shaanxi (2020JM-556) and the Fundamental Research Funds for the Central Universities (3102019ghxm016). 
Y. Li and C. Shen are the corresponding authors.
}
}

\author{Haokui Zhang,
 Ying Li,
 Hao Chen,
 Chengrong Gong,
 Zongwen Bai,
 Chunhua Shen
}

\institute{
           HZ, YL, CG\at
           $ ^1$Northwestern Polytechnical University, China
           \and
           HZ\at
           $^2$Intellifusion, China
           \and 
           ZB \at$ ^3$Shaanxi Key Laboratory of Intelligent Processing for Big Energy Data, Yan’an University, China
}

\date{\today}

\maketitle

\begin{abstract}

Recently, much attention has been spent on neural architecture search (NAS),  %
aiming  to 
outperform those manually-designed neural architectures on high-level vision recognition 
tasks. Inspired by the success, here we attempt to leverage NAS techniques to automatically design efficient network architectures for low-level image restoration tasks. In particular, we propose a memory-efficient hierarchical NAS (termed \OurMethod) and apply it to two such tasks: image denoising and image super-resolution. \OurMethod adopts gradient based search strategies and b\-u\-il\-d\-s a flexible hierarchical search space, including the inner search space and outer search space. They are in charge of designing cell architectures and deciding cell widths, respectively. For the inner search space, we propose a layer-wise arch\-itec\-ture sharing s\-tr\-a\-t\-e\-gy (L\-W\-A\-S), resulting in more flexible arch\-itectures and better performance. For the outer search space, we %
design 
a cell-sharing strategy to save memory, and considerably accelerate  the search speed. The proposed \OurMethod 
method 
is both memory and computation efficient. With a {single GTX\-1080\-Ti} GPU, it takes only about \textit{1 hour} for searching for denoising network on the BSD-500 dataset and \textit{3.5 hours} for searching for the super-resolution structure on the DIV2K dataset. 
Experiments
show that the architectures found by \OurMethod have fewer parameters and enjoy a faster inference speed, while achieving highly competitive performance compared with state-of-the-art methods. Code is available at:

\url{https://github.com/hkzhang91/HiNAS}


\end{abstract}

\tableofcontents
\clearpage

\section{Introduction}
\label{intro}
As a classical task in computer vision, image restoration aims to estimate the underlying image from its degraded measurements, which is known as an ill-posed inverse procedure. Depending on the type of degradation, image restoration can be categorized into various sub-problems, \textit{e.g.}, denoising, super-resolution, deblur,  dehaze, and inpainting, \textit{etc}.  In this work, we focus on image denoising and image super-resolution.  

Image denoising aims to restore a clean image from a noisy one. Owing to the fact that noise corruption always occurs in the image sensing process and 
inevitably 
degrades the visual quality of collected images, image denoising is needed for various computer vision tasks \citep{chatterjee2009denoising}. The purpose of single image super-resolution is to estimate a higher-resolution image from a low-resolution one. In early days, single image sup\-er-re\-so\-lu\-ti\-on methods mainly are  based on interpolation, such as nearest-neighbor, bilinear and bicubic. These simple methods are computationally very efficient, but do not yield satisfactory results. Recent %
methods 
on image denoising and image super-resolution have shifted their approaches to deep learning,
which build a mapping function from low-quality images to the desired corresponding high-quality images using convolutional networks or Transformers, and have often outperformed conventional methods significantly \citep{mao2016image, tai2017memnet, Tobias2018Neural,liu2019dual, dong2015image, zhang2018learning, zhang2018residual, dai2019second}. To date, most of these methods focus on improving the quality of the restored images, and largely neglect the inference speed.  As such, these image restoration models typically contain millions or even tens of millions of parameters. Some methods involve a recurrent optimization process to improve restoration quality, resulting in slow inference speed. Effort has 
been 
spent on employing compact models for faster inference. However, it is not trivial to manually design compact models that enjoy both good %
accuracy 
and a fast inference speed.

Recently, a growing interest is witnessed in developing solutions to automate the manual process of architecture design. Architectures automatically found by algorithms have achieved highly competitive performance in high-level vision tasks such as image classification \citep{zoph2016neural}, object detection \citep{ghiasi2019fpn, NASFCOS} and semantic segmentation \citep{liu2019auto,nekrasov2019fast}. Very recently, several works on using NAS algorithms to design architectures for image restoration have been proposed. For instance, FALSR \citep{chu2019fast}, E-CAE \citep{suganuma2018exploiting} and EvoNet \citep{liu2019deep}. Compared with the manually designed architectures, the architectures found by NAS algorithms achieve higher performance and/or have fewer parameters. A drawback is that the search process of these methods are very computationally demanding. 
By using four Tesla V100 GPUs, the method of E-CAE %
needs 
44 hours on searching. FALSR takes about 3 days on 8 Tesla V100 GPUs to find promising architectures.

In summary, deep learning based image restoration methods show promising performances, but designing an efficient architecture requires substantial efforts. Current %
NAS based image restoration approaches overcome the problem of designing architectures, but introduce a new problem---these methods require a relatively large number of computing resources. In addition, most of these methods %
do not pay attention to 
the inference speed, which is very important in practice. In this paper, we attempt to solve these problems by designing a memory-efficient NAS algorithm. Specifically, we propose a memory-efficient NAS algorithm. The proposed NAS algorithm can automatically search for neural architectures \textit{efficiently} for low-level image processing tasks, which solves the problem of designing architectures requiring substantial efforts. As the proposed is memory-efficient, it does not need %
a significant amount of 
computing resources. In addition, we also take the inference speed into consideration.

In \citep{zhang2020memory}, we have used deformable convolution to build the search space, as it is more flexible and powerful than conventional convolutions. The searched results also prove that deformable con\-vo\-lu\-tion is very useful in improving performance. %
However, 
we later find that deformable convolutions %
can be more 
time-consuming than standard convolutions. Thus, in this paper, deformable convolution is abandoned to improve the inference speed. Unfortunately, although  abandoning deformable convolution does improve the inference speed,
it %
leads to  
lower %
accuracy. 
To fill this gap, we design a new search space, by proposing a layer-wise architecture sharing strategy and adopting the residual learning structure. Benefiting from this, the architectures founded by %
our method here
enjoys a 
faster inference speed, while %
achieving 
good performance.

We %
demonstrate 
the effectiveness of our method on two %
image processing 
applications, namely image denoising and image super-resolution. Our main contributions can be summarized as follows.
\begin{enumerate}

\item Built %
upon 
gradient based NAS algorithms, here we propose a memory- and co\-m\-pu\-ta\-ti\-on-e\-ff\-i\-ci\-en\-t hierarchical neural architecture search approach for image denoising and image super-resolution, termed \OurMethod. To our knowledge, this may be the first attempt to apply differentiable architecture search algorithms to low-level vision tasks.

\item We propose a layer-wise architecture sharing strategy to improve the flexibility of the search space and propose cell sharing to save memory. Both strategies contribute to the efficiency of the proposed \OurMethod, which only takes hours to  search for architectures for image restoration tasks with \textit{a single GTX 1080Ti GPU. }

\item We apply \OurMethod to image denoising with various noise modes, and image super-resolution for evaluation. Experiments show that the networks found by our \OurMethod achieves highly competitive performance compared with state-of-the-art algorithms, while having fewer parameters and faster inference speed. 

\item We conduct experiments to analyze the network architectures found by our NAS algorithm in terms of the internal structure, offering some insights in architectures found by NAS. 

\end{enumerate}

We review some relevant work next.

\section{Related Work}

\subsection{Image Denoising and Super-resolution}

Currently, due to the popularity of convolutional neural networks (CNNs), image restoration algorithms including image denoising  and image super-resolution achieved a significant performance boost. For image denoising, the notable network models, DnCNN \citep{zhang2017beyond} and IrCNN \citep{zhang2017learning}, predict the residue, instead of the denoised image. FFDNet \citep{zhang2018ffdnet} attempts to address spatially varying noise by appending noise level maps to the input of DnCNN. NLRN \citep{liu2018non} incorporates non-local operations into a recurrent neural network (RNN) for image restoration. N3Net \citep{Tobias2018Neural} formulates a differentiable version of nearest neighbor search to further improve DnCNN.

Recently, some algorithms focus on using real-world noisy-clean image pairs to train the deep denoising model. Previously,  almost all such models are trained using synthetic noisy-clean image pairs, and there is a risk of overfitting the model to the synthetic data. Along this line, CBDNet \citep{guo2019toward} uses a simulated camera pipeline to 
generate more realistic training data. Similar work in \citep{jaroensri2019generating} proposes a camera simulator that aims to accurately simulate the degradation and noise transformation performed by the image sensing pipeline of a camera.

For image super-resolution, the first attempt was proposed in \citep{dong2015image}, where Dong \textit{et al.}\  built a CNN model which consists of three convolutional layers. Compared with traditional approaches, this neural network model 
achieved impressive performance. Later, consistent with the development of CNN %
for other vision tasks, 
CNN based super-resolution approaches %
tend 
to employ deeper %
models. For instance, Kim \textit{et al.}\ proposed VDSR \citep{kim2016accurate} and DRCN \citep{kim2016deeply}. Both VDSR and DRCN contain 20 convolutional layers and employ the residual learning strategy. To address the issue of lacking of long-term memory (one state is usually influenced by a specific prior state). Tai \textit{et al}.\  \citep{tai2017memnet} designed a memory block based on the recursive unit and gate unit, then built a MemNet by stacking the proposed block and connect them with skip connections.

In addition to designing deeper networks to improve accuracy, %
a few methods 
take spatial correlations into consideration and researchers attempt to embed the attention mechanism in their networks. For example, NLRN \citep{liu2018non} incorporates non-local modules in a recurrent network.
Based on SENet \citep{hu2018squeeze}, Zhang \textit{et al.}\  \citep{zhang2018image} built a very deep residual channel attention network for s\-u\-p\-er-r\-e\-s\-o\-lu\-ti\-on.  Recently, Dai \textit{et al.}\  \citep{dai2019second} further extended the first-order channel attention to second-order channel attention. Liu \textit{et al.}\  \citep{liu2020residual} group several residual modules together via skip connections to aggregate features.

Apart from the image restoration quality, inference efficiency is %
critically 
important for image restoration tasks. Image restoration algorithms which can restore high-quality images from low-quality images in real time would be highly valuable. Based on light-weight network 
design, Ahn \textit{et al.}\  \citep{ahn2018fast} proposed a cascading residual network and Hui \textit{et al.} \citep{hui2019lightweight} combined light-weight designing with information 
distillation. Very recently, Lee \textit{et al.} \citep{lee2020learning} boost the inference speed of FSRCNN \citep{hui2018fast} by using knowledge distillation. There is typically a trade-off between the accuracy and inference speed.

\subsection{Network Architecture Search (NAS)}

NAS aims to design automated  approaches  for discovering high-performance neural architectures such that the procedure of tedious and heuristic manual design of neural architectures can be largely eliminated from the pipeline. 

As the optimization objective of NAS is very complex and in general non-differentiable, 
early approaches often employ evolutionary algorithms (EA) for optimizing the neural architectures and parameters. The best architecture may be obtained by iteratively mutating a population of candidate architectures \citep{liu2017hierarchical}. An alternative to EA is to use reinforcement learning (RL) techniques, \textit{e.g.}, policy gradient \citep{zoph2018learning,NASFCOS} and Q-learning \citep{zhong2018practical}.
With RL, one trains a recurrent neural network that acts as a meta-controller to generate
potential ar\-ch\-i\-te\-c\-t\-ure\-s---%
typically encoded as sequences---%
by exploring a predefined search space. Note that EA and RL based methods suffer from a common drawback---inefficient  search---often requiring a large amount of computation, as these methods can be viewed as zero-order optimization.

Speed-up techniques are therefore proposed to remedy this issue. Exemplar works include hyper-networks \citep{zhang2018graph}, network morphism  \citep{elsken2018efficient} and shared weights \citep{pham2018efficient}. Recently, an increasing number of NAS %
methods 
search for architecture via (first-order) gradient-based optimization \citep{liu2018darts, cai2019device, liu2019auto}. They relax the architecture representation as a supernet via continuous relaxation, then optimize architecture parameters of this supernet via %
gradient descent.
Here, our  proposed method adopts gradient-based search strategy. 

Besides search strategies, 
how to design the 
search space %
is 
also very important for NAS algorithms. The search space defines which architectures can be represented. Early NAS approaches employ a very complex space of ch\-ai\-n-s\-tr\-u\-c\-t\-ur\-e\-d search space. This ch\-ai\-n-s\-tr\-u\-c\-t\-ur\-e\-d search space can be represented as a sequence of $n$ layers, each of which has different candidate operations, including different choices of convolution operations (depth-wise separable convolutions, dilated convolution) and pooling operations, where convolution operations can have different hyperparameters (the number of kernels, kernel sizes, and strides). Conducting architecture search on such 
a complex 
search space is computationally expensive, as the search space covers a wide range of architectures. For instance, \citet{zoph2016neural} use a few hundred GPUs and take several days to carry out architecture search experiments on Cifar-10. Later, motivated by the fact that hand-crafted architectures usually consist of repeated modules, \citet{he2016identity}
and \citet{zoph2018learning} proposed 
the 
c\-e\-ll-s\-tr\-u\-c\-t\-u\-re  search space. 
Different 
from 
the 
c\-h\-ai\-n-s\-tr\-u\-c\-tu\-re\-d search space 
where the overall 
architecture is searched  directly, the cell-structure search space 
searches 
for an architectural building block and then construct the whole network by stacking the building block following a pre-designed outline structure. Thus, 
compared with the chain-structured search space, the cell-structured search space %
offers
two major advantages: 1) the cell-structured search space covers a part of the model as cells, which consists of fewer layers, %
resulting in a 
much faster search speed. By adopting the cell-structured search space, \citet{zoph2018learning} achieve a  speed-up
of 7$ \times$
compared with %
\citep{zoph2016neural}; 2) The found architectures can be more easily transferred to other datasets by increasing or reducing the number of cells. Therefore, the cell-structured search space gradually gains popularity in recent works \citep{cai2018efficient, pham2018efficient, zhong2018practical, liu2018darts}.

The search space of our proposed method integrates
both the 
ch\-ai\-n-s\-tr\-u\-c\-tur\-ed search space and c\-e\-ll-s\-tr\-u\-c\-t\-u\-r\-ed space,
by
proposing a layer-wise architecture strategy. In previous works, the search algorithm only searches for a single cell and then constructs the %
overall 
network by stacking this cell repeatedly. Different layers have the same architecture, which %
sacrifices  some flexibility. 
In this paper, for a network consisting of a sequence of $L$ cells, the proposed search algorithm searches 
for
$L$ different cells. Designing different architectures for different layers is more flexible, and this design idea is consistent with that in Inception series~\citep{2015Going, 2016Rethinking, 2016Inception}. Benefiting from the fact that 
our search algorithm is gradient based, designing different architectures for different layers does not introduce any additional computational cost but $L-1$ sets of continuous variables. In other words, the proposed method enjoys the fast search speed of the cell-structured search space on the one hand, and has a more flexible ``ch\-ai\-n-s\-tr\-u\-c\-t\-ur\-ed-li\-ke" search space on the other hand. 

Our work is most closely related to DARTS \citep{liu2018darts}, ProxylessNAS \citep{cai2019device}, DenseNAS \citep{2020Densely}, and Auto-Deeplab \citep{liu2019auto}. All these methods take the architecture searching process as an optimization process. They first build a supernet, which consists of all possible layer types and
the 
corresponding weight parameters as the search space. Then they optimize the parameters of this supernet via %
gradient descent. %
Finally, they derive the final architecture according to weight parameters. DARTS firstly proposed the continuous relaxation of the architecture representation, allowing the efficient search of the cell architecture using gradient descent, which has achieved competitive performance. Motivated by this search efficiency, following DARTS  and its successors, \OurMethod also employs the gradient-based approach as its search strategy. Different from DARTS, the later four NAS algorithms include the kernel widths into their search space. In addition, the later four NAS algorithms discard the operation of searching for reduction cells. For ProxylessNAS, DenseNAS and Auto-Deeplab, the adjustment of spatial resolution is integrated into their candidate operations. For \OurMethod, both the reduction cell and pooling operations are discarded to retain the high resolution of feature maps. The adjustment of the receptive field relies on selecting operations of different receptive fields. ProxylessNAS discovers sequential structures and chooses kernel widths within manually designed blocks (Inverted Bottlenecks \citep{he2016identity}). DenseNAS extends the chain-structured search space and inserts more blocks %
of 
various widths %
at 
each stage. Each block is densely connected to its subsequent ones.

Different  
from ProxylessNAS and DenseNAS, the main body of \OurMethod is cell-structured and the final structure is cell-based. Both Auto-Deeplab and \OurMethod are 
designed for 
dense prediction tasks. By introducing multiple paths of different widths, Auto-Deeplab extends its search space by including kernel widths. The search space of our proposed \OurMethod resembles Auto-Deeplab.  The three main differences are as follows.
\begin{itemize}
    \item 
For candidate operations, %
we discard pooling operations to retain the high resolution of feature maps and rely on automatically selected operations of different receptive fields to adapt the receptive field; 
\item 
We propose a layer-wise architecture-sharing strategy to improve flexibility; 
\item 
A cell-sharing strategy is proposed for improving memory efficiency.

\end{itemize}

Relevant to using NAS algorithms to search for neural network architectures for low-level image restoration tasks, three most related works are EvoNet \citep{liu2019deep}, E-CAE \citep{suganuma2018exploiting} and FALSR \citep{chu2019fast}. EvoNet searches for networks for medical image denoising via EA. E-CAE \citep{suganuma2018exploiting} employs EA to search for an architecture of convolutional auto-encoders for image inpainting and denoising. FALSR was proposed for image super-resolution tasks. FALSR combines RL and EA and designs a hybrid controller as its model generator. All three methods mentioned above require a relatively large a\-m\-o\-u\-n\-t of computations and take a large amount of GPU time for searching. Compared with these methods, our proposed \OurMethod employs a different search space, and search strategies, leading to significantly higher search efficiency.

We now present our method in detail.

\section{Our Method}
\label{sec: our_method}

Our proposed \OurMethod is a gradient-based architecture search algorithm. Specifically, we search for $L$ different computation cells, where $L$ denotes the number of cells. The final architecture is built  by stacking the $L$ cells of different widths one by one. To be able to search for both cell topological architectures and cell widths, \OurMethod builds a hierarchical search space. The inner space is responsible for searching for the inner cell topological architectures, and the outer space is in charge of searching for cell widths. To further improve the efficiency of \OurMethod, we propose a cell-sharing strategy, allowing features from different levels of the outer search space to share one cell. 

In this section, we first introduce how to search for architectures of cells based on continuous relaxation (inner search space). Then we explain how to determine the widths via multiple candidate paths and cell sharing (outer search space). Then, we elaborate on residual learning frameworks for image denoising and super-resolution. Last, we present our search strategy and details of the loss functions.

\subsection{Inner Search Space}

\begin{figure*}
\centering 
\includegraphics[width=0.55\textwidth]{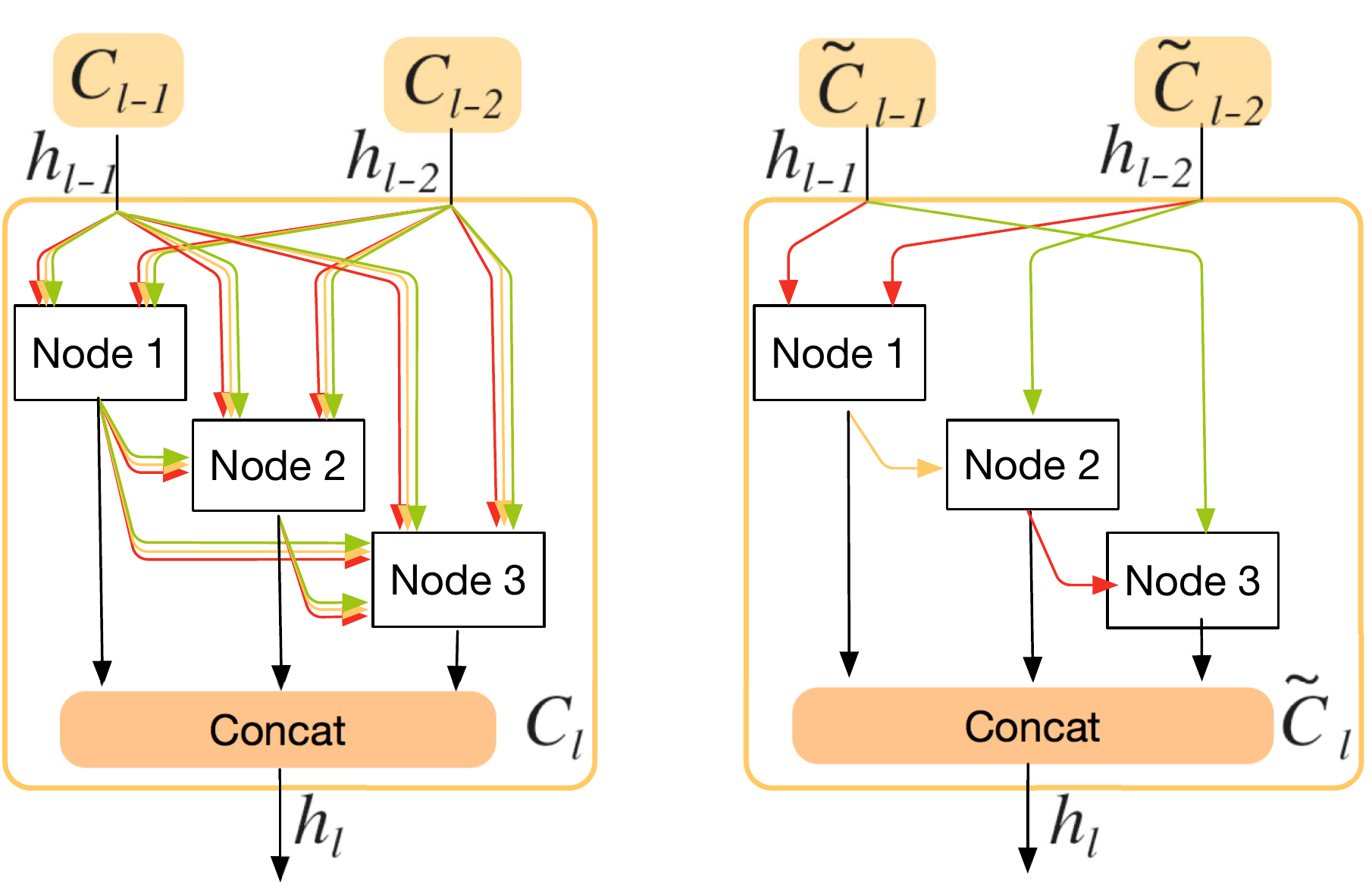}
\caption{Inner-cell architecture search. Left: the supercell that contains all possible layer types. Right: the cell architecture searching result, a compact cell, where each node only keeps the two most important inputs and each input is connected to the current node with a selected operation.}
\label{fig:cell_architecture_search} 
\end{figure*}

\begin{figure*}[h]
\includegraphics[width=1.0\textwidth]{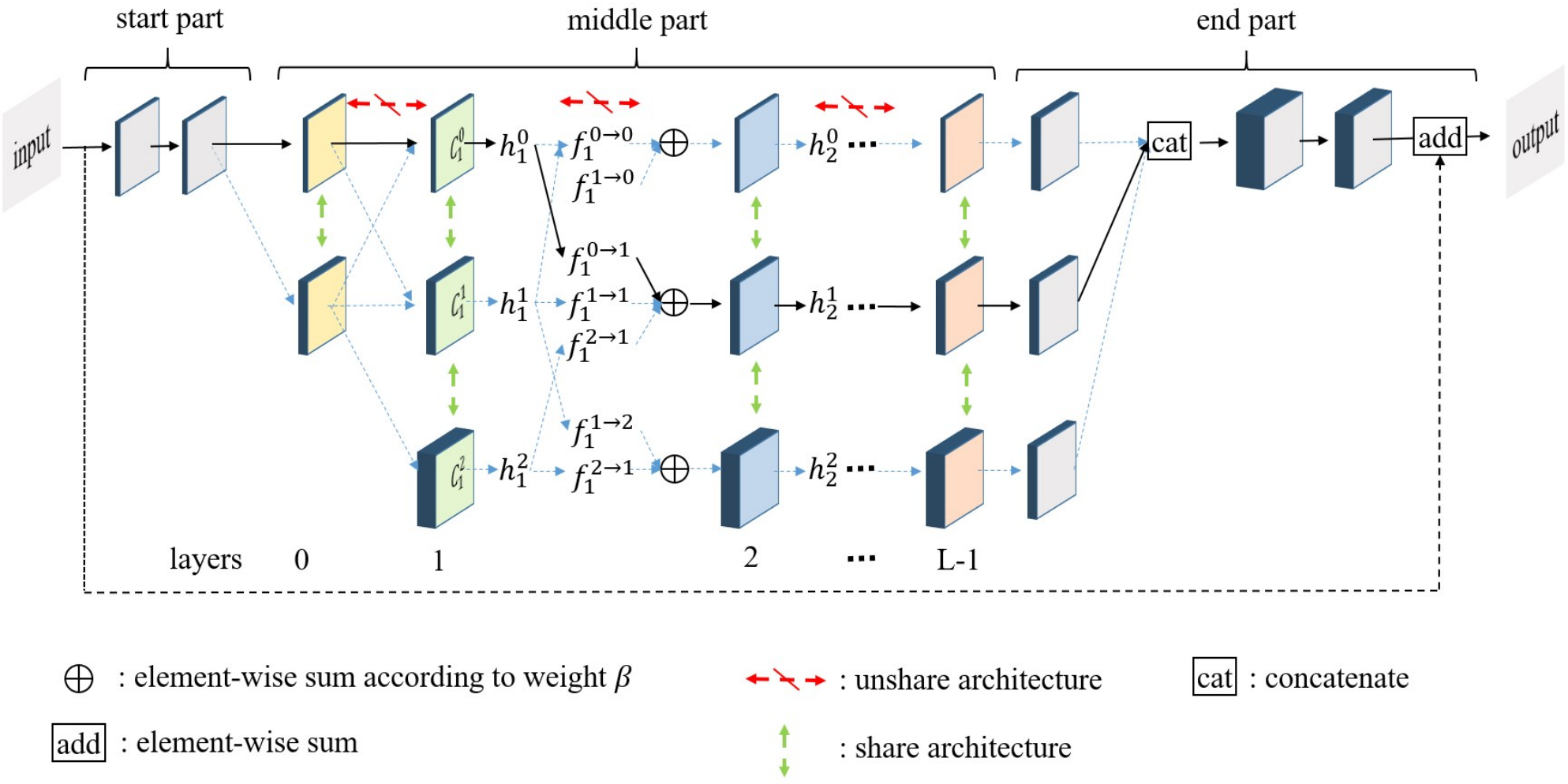}
\caption{The overall framework of the supernet. The supernet contains three components, including the `start part', `middle part', and `end part'. The start and end parts are manually pre-designed and their architectures are fixed during search. The middle part is %
automatically found 
by the search algorithm. It consists of $L$ layers, each of which contains three cells of different widths. In the proposed \OurMethod, cells in the same layer share the same architecture and cells in different layers can have 
different architectures. This improves the flexibility of the search space, resulting in better performance, while introducing no computation overhead. During search, all paths are activated, and the weight $\beta$ is updated via gradient descent. At the end of search, each layer keeps only one path, which is marked in a black arrow.}
\label{fig:network_architecture_search} 
\end{figure*}

To search for the inner cell architecture, we first build a super-cell containing $N$ nodes. Different nodes in this supercell are connected with different paths, and each path contains all possible operations. Each operation in each path corresponds to a weight $\alpha$, which denotes the importance of this layer type in the current path. The purpose of the cell architecture search is to learn  a set of continuous variables $\{\alpha\}$. During search, continuous variables are updated via gradient descent optimization. At the end of the search, a discrete architecture can be obtained by keeping the top two %
\textit{most probable }
layer types and discarding the rest for each node. Next, we explain this process in detail.

\vspace{5 pt}
\noindent {\bf Shared inner-cell architecture search}. We denote the supercell in layer $l$ as ${C}_{l}$, which takes outputs of previous cells and the cell before previous cells  as inputs and outputs a tensor ${h}_{l}$, 
as shown in Figure \ref{fig:cell_architecture_search}.
The left-side is the supercell containing all possible operations. It is a directed acyclic graph consists of three nodes. Inside ${C}_{l}$, each node takes the two inputs of the current cell and the outputs of all previous nodes as input and outputs a tensor ${h}_{l}$. When all layers share the same cell architecture, the output of the $i$th node in the cell is calculated as:
\begin{equation}
\begin{split}
{}& {x}_{l,i}=\sum_{{x}_{j}\in{I}_{l,i}}^{}{{O}^{j\rightarrow i}({x}_{j})},\\
{}& { O }^{ j\rightarrow i }({ x }_{ j })=\sum _{ k=1 }^{ S }{ { { \alpha  }_{k}^{j\rightarrow i } }{ o }^{ k } } ({ x }_{ j }),
\end{split}
\end{equation}
where ${I}_{l,i}=\{{h}_{l-1}, {h}_{l-2}, {x}_{l,j<i}\}$ is the input set of node $i$. ${h}_{l-1}$ and ${h}_{l-2}$ are the outputs of cells in layers $l-1$ and $l-2$, respectively. ${O}_{l}^{j\rightarrow i}$ is the set of possible layer types. $\{{o}^{1}, {o}^{2}, \cdots, {o}^{S}\}$ correspond to $S$ possible operations. ${\alpha}_{k}^{j\rightarrow i}$ denotes the weight of operator ${o}^{k}$.

\vspace{5 pt}
\noindent {\bf Layer-wise architecture sharing (LWAS)}. In fact, in
early days, 
NAS methods 
often 
employ a very complex search space to search 
for 
the %
entire 
network \textit{directly}, requiring an 
enormous  
number of GPUs and taking a %
significant amount of computation.
For instance, Zoph \textit{et al.}\ used a few hundred GPUs and train for several days to search for architectures on the small dataset of Cifar-10 \citep{zoph2016neural}. Later, Zoph \textit{et al.}\  proposed the NASNet, which searches for an architectural building block then build the 
overall 
network by stacking the found block with a pre-designed outline structure \citep{zoph2018learning}. Compared with the method in \citep{zoph2016neural},  NASNet enjoys a much faster search speed. Thus, the search space design of NASNet is widely used in most recent works. NASNet significantly reduces the %
complexity 
of the search space, which is very useful for improving the search speed in RL and EA based NAS algorithms. Reducing the complexity of search space means training fewer student networks. The NASNet search space accelerates the search speed on one hand, with the price of %
restricting 
the flexibility of the search space.
It restricts different layers of networks to share the same cell architecture. In our proposed \OurMethod, we adopt a more flexible search space: we %
search 
for different cell architectures, for different layers. In our gradient based \OurMethod, we only need to introduce more continuous variables ${\alpha}$ without any additional computational cost. In our new search space, the output of the $i$-th node is computed with:
\begin{equation}
\begin{split}
{}& {x}_{l,i}=\sum_{{x}_{j}\in{I}_{l,i}}^{}{{O}_{l}^{j\rightarrow i}({x}_{j})},\\
{}& { O }_{l}^{ j\rightarrow i }({ x }_{ j })=\sum _{ k=1 }^{ S }{ { { \alpha  }_{l}^{ k,  j\rightarrow i } }{ o }^{ k } } ({ x }_{ j }).
\end{split}
\end{equation}
Here, ${\alpha}_{l}^{k
, j\rightarrow i}$ is $l$ related. The sets of continuous variables ${\alpha}$ in different layers are different. Cells in the same layer share the same set of continuous variables ${\alpha}$. ${h}_{l}$ is the concatenation of the outputs of $N$ nodes, and it can be expressed as:
\begin{equation}
\begin{split}
{h}_{l}={}& {\rm Cell} ({h}_{l-1}, {h}_{l-2})\\
={}& {\rm Concat} \{{x}_{l,i}|i\in\{1, 2, \cdots, N\}\}. 
\end{split}
\end{equation}

\vspace{5 pt}
\noindent {\bf Search space}. In this paper, we employ the following 7 types of operators: 
\begin{itemize}
\itemsep 0pt 
    \item {\tt conv}: $3\times 3$ convolution;
    \item {\tt sep}: $3\times 3$ separable convolution;
    \item {\tt sep}: $5\times 5$ separable convolution;
    \item {\tt dil}: $3\times 3$ convolution with dilation rate of 2;
    \item {\tt dil}: $5\times 5$ convolution with dilation rate of 2;
    \item {\tt skip}: skip connection;
    \item {\tt none}: no connection and return zero.
\end{itemize}

To preserve pixel-level information for low-level image processing, we abandon down-sample operations such as pooling layers;
and 
set the 
stride to 2 in convolution layers. For convolution operators, we employ three convolution types including %
standard 
convolutions, separable convolutions and dilation convolutions. Each convolution operator, in order, consists of a LeakyReLU activation layer, a convolution layer and a batch normalization layer. We provide two different kernel sizes of 3 and 5 for separable and dilation convolutions. The search algorithm can adapt the receptive field by selecting convolution operators 
of 
different kernel sizes. An example of found compact cell is shown in the right-side of Figure~\ref{fig:cell_architecture_search}.

\begin{figure*}[h!]
\begin{center}
\includegraphics[width=.6\textwidth]{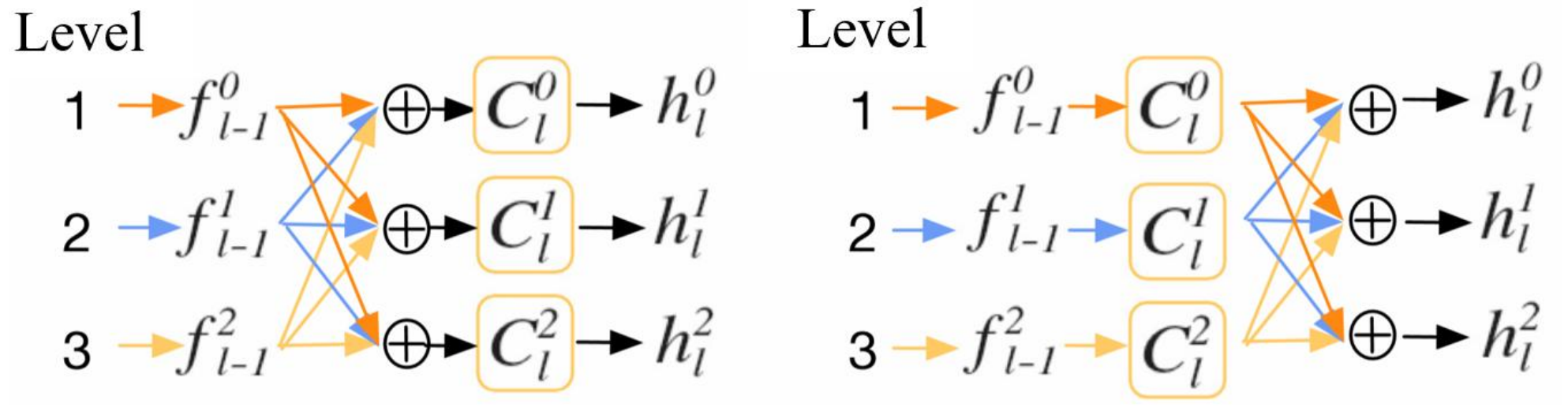}
\end{center}
\caption{Comparison of cases of whether using cell sharing or not. Left: features from different levels share the same cell. Using cell sharing; Right: features from different levels use different cells.}
\label{fig:cell sharing}
\end{figure*}

\subsection{Outer Search Space}

Now, the main idea of searching for specific architectures inside cells has been presented. Beside search architectures 
of 
inner cells, we still need to either \textit{heuristically} set the width of each cell or \textit{automatically} search for a proper width for each cell, in order to build the overall network.

\vspace{5 pt}
\noindent {\bf Multiple candidate paths}. In conventional CNNs, the change of widths of convolution layers is often related to the change of spatial resolutions. For instance, once the features are down-sampled, the widths of following convolution layers are doubled. In our \OurMethod, instead of using down-sample operations such as pooling layers and setting the stride to 2 in convolution layers, we rely on operations with different receptive field such as dilation convolution operations of $3\times3$ and $5\times5$, and separable convolution operations of $3\times3$ and $5\times5$ to adjust the receptive field \textit{automatically}.

Thus, the conventional experience of adjusting width no longer applies to our case. To solve this problem, we employ the flexible hierarchical search space and leave the task of deciding the width of each cell to the NAS algorithm itself, making the search space more general. In fact, several NAS algorithms in the literature also search for the outer layer width, mostly for high-level image understanding tasks. For example, FBNet \citep{wu2019fbnet} and MNASNet \citep{tan2019mnasnet} consider different expansion rates inside their modules to discover compact networks for image classification.  

In this section, we introduce the outer layer width search space, which determines the widths of cells in different layers. Similarly, we build a supernet that contains several supercells with different widths in each layer. As illustrated in Figure~\ref{fig:network_architecture_search}, the supernet mainly consists of three parts: 
\begin{enumerate}
    \item 

\textit{`start-part'}, consisting  of an 
input layer and two convolution layers. A relatively shallow feature is extracted by feeding the input data to two convolution layers, and a copy of input data is propagated to the `\textit{end-part}' via skip connections; 

\item 
\textit{`middle-part'}, containing $L$ layers and each layer having three supercells of different widths. This is 
the main part of supernet. The shallow feature extracted in 
the `start-part' is fed to different cells of layer $0$.
Then the outputs of layer $0$ are fed to different cells of layer $1$, and so on;

\item 
\textit{`end-part'}, concatenating the outputs of layer $L-1$, then feeding them to two convolution layers to generate the residual, finally element-wise summing the learned residual and the output of skip connection to %
obtain 
the final result. 
\end{enumerate}

Except layer $0$, which contains two cells of different widths, our supernet provides three paths of cells with different widths, which correspond to three decisions: 1) reducing the width; 2) keeping the previous width; 3) increasing the width. After searching, only one cell at each layer is kept.  
The continuous relaxation strategy mentioned in the cell architecture search section is reused %
here.

At each layer $l$, there are three cells ${C}_{l}^{0}$, ${C}_{l}^{1}$ and ${C}_{l}^{2}$ with widths ${\gamma}^{0}\times W$, ${\gamma}^{1}\times W$ and ${\gamma}^{2}\times W$, where $W$ is the basic width and ${\gamma}^{0}$, ${\gamma}^{1}$ and ${\gamma}^{2}$ are width changing factors. The output feature of each layer is
\begin{equation}
{ h }_{ l}=\{{h}_{l}^{0}, {h}_{l}^{1}, {h}_{l}^{2}\},
\end{equation}
where ${h}_{l}^i$ is the output of ${C}_{l}^i$. The channel width of ${h}_{l}^{i}$ is ${2}^{i}NW$, where $N$ is the number of nodes in the cells.

\begin{figure*}
\centering 
\includegraphics[width=0.7065\textwidth]{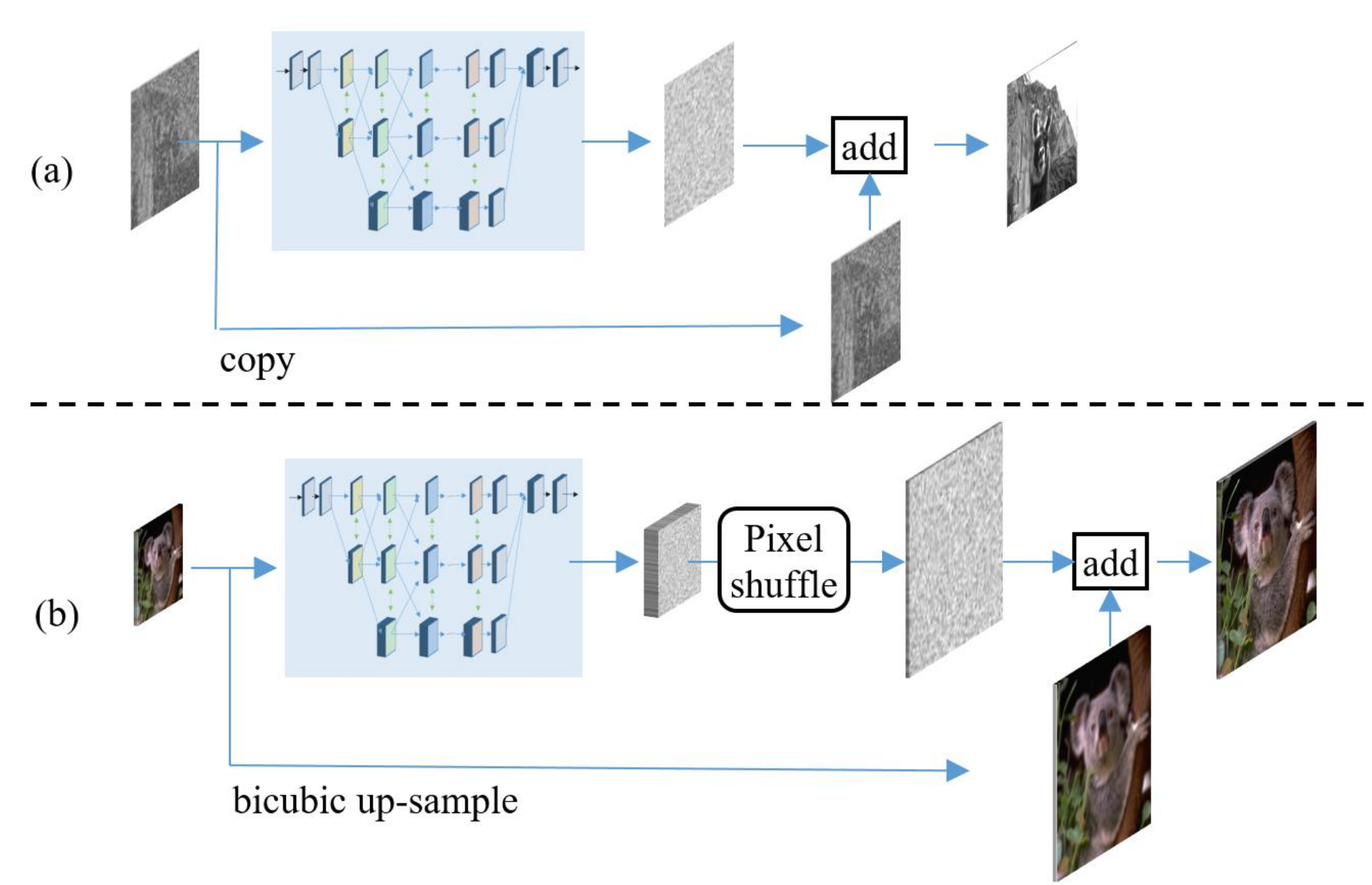}
\caption{Residual learning frameworks. (a) residual learning framework for denoising; (b) residual learning framework for super-resolution. }
\label{fig:residual_learning} 
\end{figure*}

\vspace{5 pt}
\noindent {\bf Cell sharing}. Each cell $C_l^i$ is connected to $C_{l-1}^{i-1}$, $C_{l-1}^{i}$ and $C_{l-1}^{i+1}$ in the  previous layer and $C_{l-2}^{i}$ two layers before. We first process the outputs $h_{l-1}$ from those layers with a $1\times1$ convolution to form features $f_{l-1}$ with width $2^iW$,  matching the input of $C_l^i$. Then the output for the $i$th cell in layer $l$ is computed with
\begin{equation}
{h}_{l}^{i}=C_l^i \left(
\sum_{k=i-1}^{i+1}{\beta}_{k}^{i}{f}_{l-1}^{k},{f}_{l-2}^{i}
\right),
\label{formula: cell sharing}
\end{equation}
where ${\beta}_{k}^{i}$ is the weight of ${f}_{l-1}^{k}$. We combine the three outputs of ${C}_{l-1}$ according to corresponding weights, then feed them to ${C}_{l}^{i}$ as input. Here, features ${f}_{l-1}^{i-1}$, ${f}_{l-1}^{i}$ and ${f}_{l-1}^{i+1}$ come from different levels, but they share the cell ${C}_{l}^{i}$ during computing ${h}_{l}^{i}$.

Note the similarity of this design with that of 
Auto-Deeplab, which is used to select feature strides for image segmentation. However, in Auto-Deeplab, the outputs from the three different levels are first processed by separate cells with different sets of weights before summing into the output:
\begin{equation}
{h}_{l}^{i}=\sum_{k=i-1}^{i+1}{\beta}_{k}^{i}C_l^{k}({f}_{l-1}^{k},{f}_{l-2}^{i}), 
\label{formula: not share cell}
\end{equation}

A comparison between Eqs.~\eqref{formula: cell sharing} and \eqref{formula: not share cell} is shown in Figure~\ref{fig:cell sharing}, where the inputs from layer $l-1$ are not shown %
for simplicity. 

For the hierarchical structure which has three candidate paths, the cell in each candidate path is used once with Eq.~\eqref{formula: cell sharing} and it is used 
for 
three times with Eq.~\eqref{formula: not share cell}. By sharing the cell $C_l^i$, we are able to save the memory consumption by a factor of 3  in the supernet. Cell sharing has two main advantages: 1) Improving applicability. NAS in general consumes much memory and computation. Improving memory efficiency enables much broader applications. 2) Improving searching efficiency. As cell sharing saves memory consumption in the supernet, during search, we can use larger batch sizes %
speed up the convergence and search speed.
It 
also enables the use of 
a deeper and wider supernet for more accurate approximations.

As memory efficiency is important for NAS methods, some previous methods have already been proposed for better efficiency \citep{pham2018efficient, dong2019searching, cai2019device, dong2019one, guo2020single}. Compared with memory-efficient s\-tr\-a\-te\-g\-i\-e\-s  adopted in these methods, our proposed cell sharing strategy %
differs 
in the core idea. For ENAS \citep{pham2018efficient}, the core idea is loading pretrained weight into new the generated student networks to reduce the time cost of training from scratch. GDAS \citep{dong2019searching}, Proxyless-NAS \citep{cai2019device}, SETN \citep{dong2019one} and Single Path One-Shot NAS \citep{guo2020single} share the same key idea, %
which 
is to avoid traversing all the possible  operations  
during search. %
In PC-DARTS \citep{xu2021partially}, %
for 
training the superent, a subset of channels of each path 
are 
activated in each iteration. The core idea of our cell sharing is reducing duplicate operations, processing the sum of outputs of different candidate paths rather than processing each output separately. A similar idea is also adopted in a concurrent work \citep{chen2019efficient}, the key idea of which is calculating the weighted sum of kernels rather than the output features.

Outer layer widths search %
appears to be 
an extension of the inner cell search. However, 
the way to obtain 
the final widths of different layers %
is 
different from %
that of obtaining 
the final compact cell architecture in the inner cell architecture search. If we follow previous actions 
as in the inner cell search, the widths of adjacent layers in the final network may change drastically, which %
would have 
a negative impact on the efficiency, as explained in \citep{ma2018shufflenet}. To avoid this problem, we view the $\beta$ values as a probability. Then use the Viterbi decoding algorithm to select the path with the maximum probability as the final result.

Different from these %
methods 
targeting the classification task, our search space considers the interaction between multi-level features with different receptive fields,  %
so as to 
preserve high resolutions. To simultaneously consider hierarchical routing and inner cell design choices for width and operation, we build a multi-level supernet, of which
every subnet can encode more powerful multi-level features than those single path networks.

There are several previous algorithms which also include width in their search space.
AMC \citep{he2018amc}, FBNetV2 \citep{wan2020fbnetv2} and TuNAS \citep{bender2020can} all assign different widths to different operations within a sequential search space. It is worth noting that our search space is essentially different from theirs, where operation-wise width search is not trivial. 
In addition, FBNetV2 searches for widths via channel masking. TAS \citep{dong2019network} prunes channels according to a
learnable distribution, then transfers knowledge from the unpruned network to the pruned network via knowledge distillation. Both methods do not consider topology in their search space, thus being less flexible than ours.

\vspace{5 pt}
\noindent {\bf Residual learning}
In our \OurMethod, instead of learning the final restoration results directly, we learn residual information between the low-quality images and high-quality ones to further improve the performance. Figure~\ref{fig:residual_learning} shows the residual learning frameworks for image denoising and image super-resolution.

For image denoising, a copy of input data is propagated to the end of the framework directly, then added to the output of the network to generate the final restoration result, as shown in Figure~\ref{fig:residual_learning}(a).

For image super-resolution, residual learning is not
straightforward, 
as the spatial resolutions of input and output are typically  different. As shown in Figure \ref{fig:residual_learning}(b), taking a $M\times N\times 3$-sized image as input, the network generates a $M\times N \times 3S^2$-sized output.
Here, $M$, $N$ and $S$ denote the spatial resolution of the input image and upscale factor of super-resolution, respectively. Then, the pixel shuffle operation is used to convert the $M\times N \times 3S^2$-sized output to $MS\times NS \times 3$-sized residual. Finally, the learned residual is added to the upscaled input image to generate the final restoration result.

\subsection{Searching Using Gradient Descent}
\label{sec: searching using gradient descent}

The searching process in our %
\OurMethod is the optimization process of supernet.
The loss of our \OurMethod has two %
terms. 
Inspired by the fact that PSNR and SSIM~\citep{wang2004image} are the two most widely used evaluation metrics in image restoration tasks, we design a loss consisting two %
terms,
which %
correspond 
to the two evaluation metrics, respectively. Our optimization function can be %
written 
as:
\def\ssim{{\rm ssim}}
\def\DartsDn{ {\rm IRNS } }
\begin{equation}
\begin{split}
{}& {\rm loss}  = {\left\| {\rm {f}_{net}}(x)-y\right\|}_{2}^{2} + \lambda \cdot  {\rm {l}_{ssim}}({\rm {f}_{net}}(x), y),\\
{}& {\rm {l}_{ssim}}(x,y)={\log}_{10}({{\rm ssim}(x,y)}^{-1}),\\
{}& {\rm {f}_{net}}(x) = {\rm {f}_{res}}(x) + {\rm {f}_{skip}}(x),
\end{split}
\end{equation}
where $x$ and $y$ represent the input image and corresponding ground-truth. ${\rm {l}_{ssim}}(\cdot)$ is a loss %
term 
that is designed to enforce the visible structure of the result. ${\rm {f}_{res}}(\cdot)$ is the supernet, which learns residual from input images. ${\rm {f}_{res}}(\cdot)$ is the skip connection, and it is a copy operation for image denoising and bicubic up-sample operation for image super-resolution. ${\rm ssim}(\cdot)$ is the structural similarity~\citep{wang2004image}. $\lambda$ is a weighting coefficient, and it is empirically set to 0.6 in all the experiments.

\vspace{2 pt}

During optimization of the supernet with gradient descent, we find that the performance of a network founded by \OurMethod is often observed to \textit{collapse} when the number of search epochs becomes large. The very recent method of  Darts+~\citep{liang2019darts+}, which is concurrent to this work here, presents similar observations.  
Because of this collapse issue,  it is %
challenging 
to pre-set the  number of search epochs. To %
tackle 
this %
issue, 
we keep the supernet obtaining the best performance during evaluation as the final search result. Specifically, we split the training set into three disjoint parts: Train W, Train A and Validation V. 

Sub-datasets W and A are used to optimize the weights of the supernet (kernels in convolution layers) and weights of different layer types and cells of different widths ($\alpha$ and $\beta$).  During optimizing, we periodically evaluate the performance of the trained supernet on the validation dataset V and record the best performance. After the search process is finished, we choose the supernet which corresponds to the best performance %
as the result of the architecture search. More details are presented in the search settings in Section~\ref{sec: experiments}.

\section{Experiments}
\label{sec: experiments}

\begin{figure*}[t]
\begin{center}
\includegraphics[width=6.8in]{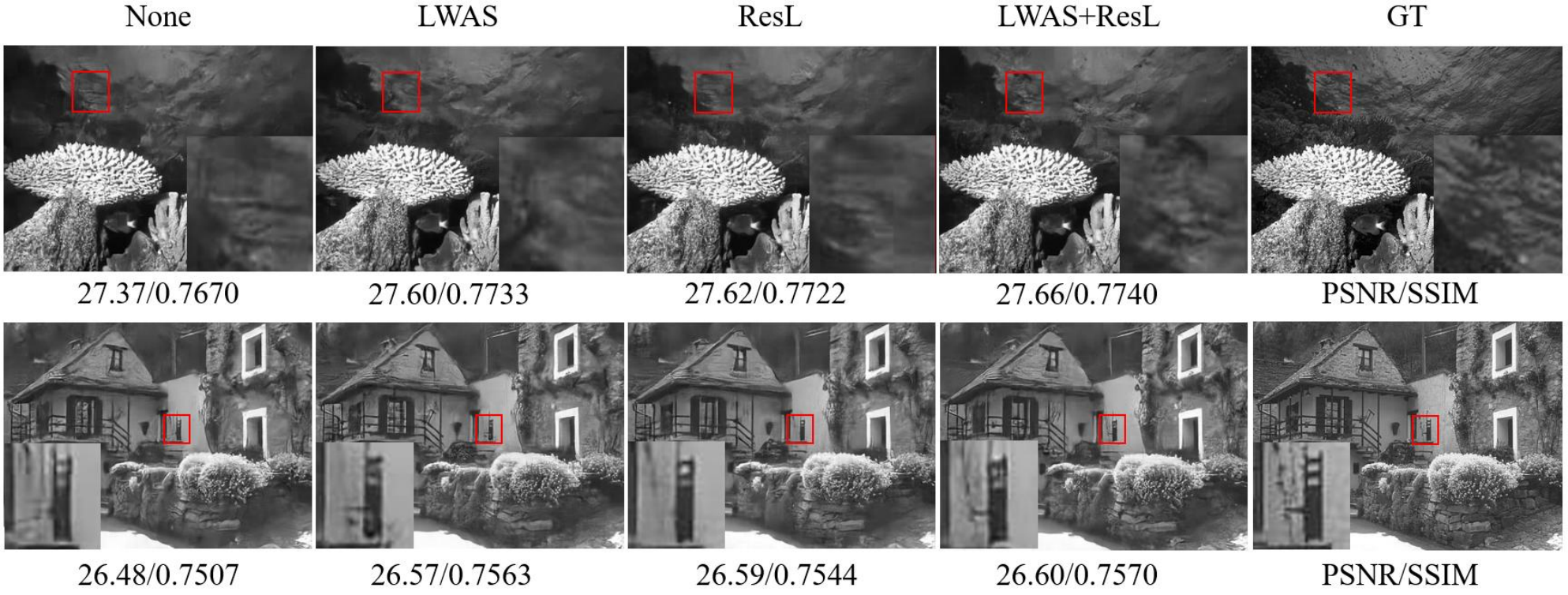}
\end{center}
\caption{Visual results of image denoising. The five columns, from left to right, are denoising results of using basic search space, using layer-wise architecture search strategy, using residual learning strategy, using both strategies and ground-truths. The top and bottom rows are the denoising results for image `101027' and `207038' from BSD200 when $\sigma=50$.}
\label{fig:ab_lwas_resl_dn}
\end{figure*}

\newcolumntype{C}[1]{>{\centering\let\newline\\\arraybackslash\hspace{0pt}}m{#1}}
\begin{table*}[h]
\caption{Effects of layer-wise architecture sharing (LWAS) and residual learning (ResL) strategies for denoising. RSWS represents random search with weight sharing \citep{li2020random}. $N$ is the number of random sampled models. We run each experiment for three times with different random seeds and report both the mean and standard deviation.}
\label{tab: ablation_denoising}
\footnotesize
\renewcommand\arraystretch{1.2}
\renewcommand\tabcolsep{6.0pt}
\begin{center}
{
\begin{tabular}{C{1.5cm}|C{1.0cm}|C{1.0cm}|C{1.8cm}|C{1.8cm}C{1.2cm}|C{1.2cm}C{1.2cm}|C{1.2cm}C{1.2cm}}
\hline
\multirow{2}*{Methods} & \multirow{2}*{LWAS}& \multirow{2}*{ResL} & Parameters & \multicolumn{2}{c|}{$\sigma=30$} & \multicolumn{2}{c|}{$\sigma=50$} & \multicolumn{2}{c}{$\sigma=70$} \\
&  &   & (M)    &  PSNR   & SSIM    & PSNR   & SSIM    & PSNR   &  SSIM    \\   
\hline

RSWS &\multirow{2}*{\ding{55}} &\multirow{2}*{\ding{55}} &\multirow{2}*{0.23-0.31} 
& 28.93        & 0.8344      & 26.60       & 0.7583      & 25.31       & 0.7003     \\ 
$N=30$ &   &   &     
& $\pm$0.1216  & $\pm$0.0023 & $\pm$0.0982 & $\pm$0.0008 & $\pm$0.0608 & $\pm$0.0025 \\

\hline
\multirow{2}*{\OurMethod} &\multirow{2}*{\ding{55}} &\multirow{2}*{\ding{55}} &\multirow{2}*{0.23-0.31} 
& 29.15        & 0.8386      & 26.70       & 0.7591      & 25.39       & 0.7024     \\ &   &   &     
& $\pm$0.0238  & $\pm$0.0022 & $\pm$0.1749 & $\pm$0.0030 & $\pm$0.0340 & $\pm$0.0023 \\

\multirow{2}*{\OurMethod} &\multirow{2}*{\ding{51}} &\multirow{2}*{\ding{55}} &\multirow{2}*{0.23-0.34}
& 29.17        & 0.8399      & 26.78       & 0.7616      & 25.42       & 0.7062     \\ &   &   &     
& $\pm$0.0096  & $\pm$0.0008 & $\pm$0.0723 & $\pm$0.0028 & $\pm$0.0126 & $\pm$0.0021 \\

\multirow{2}*{\OurMethod} &\multirow{2}*{\ding{55}} &\multirow{2}*{\ding{51}} &\multirow{2}*{0.20-0.29}
& 29.20        & 0.8407      & 26.80       & 0.7622      & 25.39       & 0.7065     \\ &   &   &     
& $\pm$0.0263  & $\pm$0.0016 & $\pm$0.0500 & $\pm$0.0014 & $\pm$0238 & $\pm$0.0010 \\

\multirow{2}*{\OurMethod} &\multirow{2}*{\ding{51}} &\multirow{2}*{\ding{51}} &\multirow{2}*{0.22-0.34}
& \bf{29.23}   & \bf{0.8411} & \bf{26.81}  & \bf{0.7649} & \bf{25.43}  & \bf{0.7069} \\ &   &   &     
& $\pm$0.0346  & $\pm$0.0015 & $\pm$0.0700 & $\pm$0.0035 & $\pm$0.0265 & $\pm$0.0006 \\

\hline
\end{tabular}
}
\end{center}
\end{table*}

\subsection{Datasets and Implementation Details}
We carry out the denoising experiments on %
the 
widely used dataset, BSD500 \citep{martin2001database}. Following \citep{mao2016image,tai2017memnet,liu2018non, liu2019dual}, we use the combination of 200 images from the training set and 100 images from the validation set as the training set, and test on 200 images from the test set. On this dataset, we generate noisy images by adding white Gaussian noises to clean images with $\sigma=30, 50, 70$. 

For image super-resolution experiments, we train on 800 high-resolution images from the DIV2K dataset and test on 5 most widely used benchmark datasets, including Set5, Set14, BSD100, Urban100 and Manga109. We carry out experiments with Bicubic degradation models. The high-resolution images are respectively degraded to 
$\nicefrac{1}{2}$, 
$\nicefrac{1}{3}$, and 
$\nicefrac{1}{4}$
of the original spatial resolution.

\vspace{5 pt}
\noindent 
\textbf{Search settings}. The supernets that we build for image denoising and image super-resolution are different. For image denoising, the supernet contains 3 cells, each of which consists of 4 nodes. We perform the architecture search on BSD500. Specifically, we randomly choose 2\% of training samples as the validation set (Validation V). The rest are equally divided into two %
subsets: 
one subset is used to update the kernels of convolution layers (Train W) and the other %
subset 
is used to optimize the parameters of the neural architecture (Train A). For image super-resolution, we build a supernet that has 2 cells and each cell involves 3 nodes. We perform the search on DIV2K. Similarly, the training samples are divided into three %
subsets,
including Validation V, Train W and Train A.

The batch size of training the supernet is set to 8 for image denoising and 24 for image super-resolution. We train the supernet for %
maximum 
100 epochs and optimize the parameters of kernels and architecture with two optimizers. For learning the parameters of convolution layers, we employ the standard SGD optimizer. The momentum and weight decay are set to 0.9 and 0.0003, respectively. The learning rate decays from 0.025 to 0.001 with the cosine annealing strategy \citep{loshchilov2016sgdr}. For learning the architecture parameters, we use the Adam optimizer, where both learning rate and weight decay are set to 0.001. In the first 20 epochs, we only update the parameters of kernels, then we start to alternately optimize the kernels of convolution layers and architecture parameters from epoch 21.

During the training process of searching, we randomly crop patches of $64\times64$ and feed them to the network. During the evaluation, we split each image into some adjacent patches of $64\times64$ and then feed them to the network and finally join the corresponding patch results to obtain the final results of the whole test image. From epoch 61, we evaluate the supernet for every epoch and update the best performance record.

\vspace{5 pt}
\noindent
\textbf{Training settings} With the SGD optimizer, we train the denoising and super-resolution networks for 400k and 600k iterations, where the initial learning rate, batch size are set to 0.05 and 24, respectively. For data augmentation, we use random crop, random rotations $\in \{{0}^{\circ}, {90}^{\circ}, {180}^{\circ}, {270}^{\circ}\}$, horizontal and vertical flipping. For random crop, the patches of $64\times 64$ are randomly cropped from input images.

\subsection{Ablation Study}

\begin{figure*}[t]
\begin{center}
\includegraphics[width=6.8in]{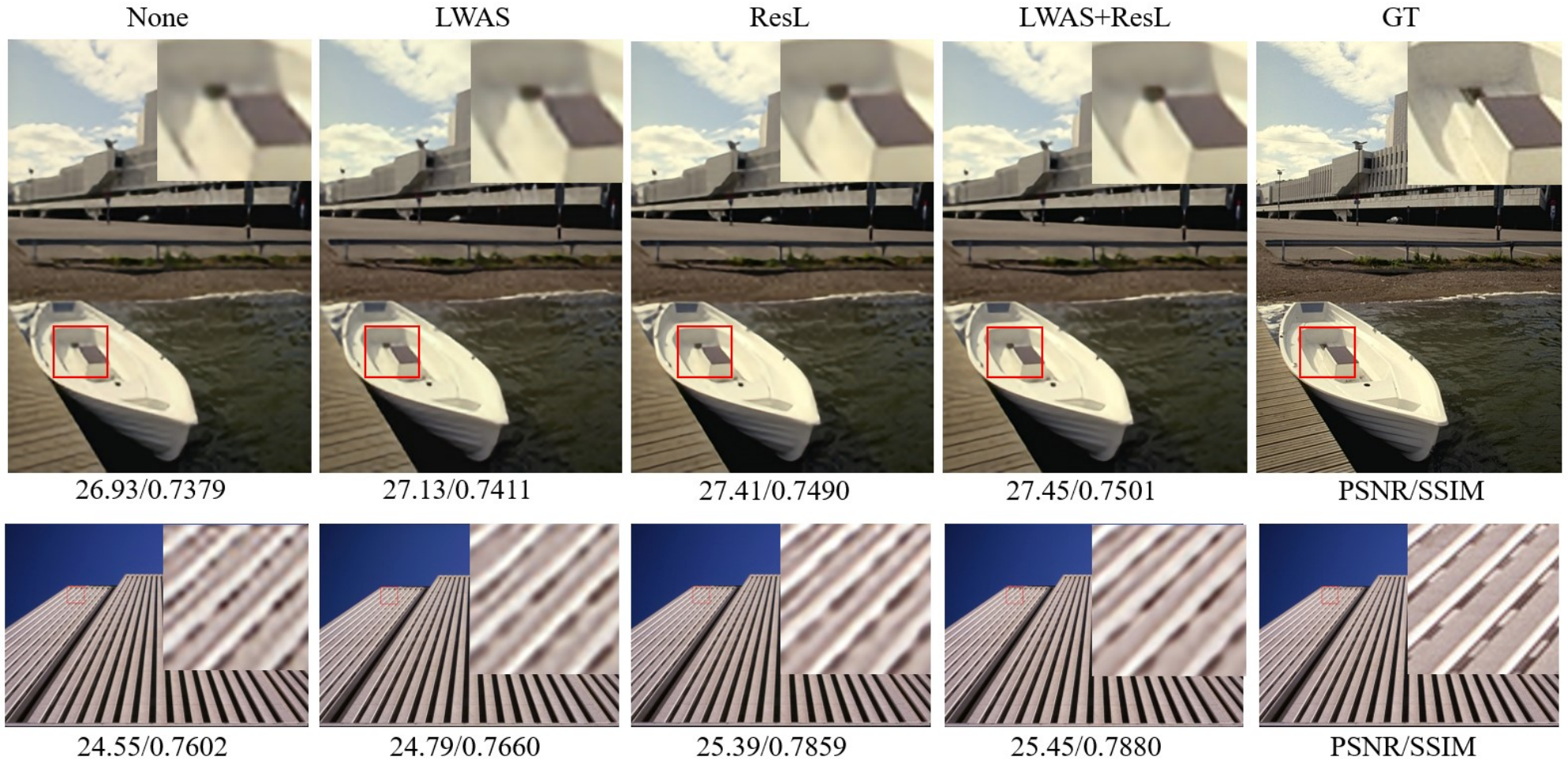}
\end{center}
\caption{Visual results of $\times$4 image super-resolution. The five columns, from left to right, are super-resolution results of using basic search space, using layer-wise architecture search strategy, using residual learning strategy, using both strategies and ground truths. The first row is the super-resolution results for image '78004' from BSD100. The second row is the super-resolution results for image '016' from Urban100.}
\label{fig:ab_lwas_resl_sr}
\end{figure*}

\begin{table*}[h]
\caption{Effects of layer-wise architecture sharing (LWAS) and residual learning (ResL) strategies for 4$\times$\ super-resolution. RSWS represents random search with weight sharing~\citep{li2020random}. $N$ is the number of random sampled models. We run each experiment for three times with different random seeds and reported both the mean and standard deviation.}
\label{tab: ablation_SR}
\footnotesize
\renewcommand\arraystretch{1.2}
\renewcommand\tabcolsep{4.50pt}
\begin{center}
{
\begin{tabular}{c|c|c|c|cc|cc|cc|cc}
\hline
\multirow{2}*{Methods}  & \multirow{2}*{LWAS} & \multirow{2}*{ResL} & Parameters & \multicolumn{2}{c|}{Set5} & \multicolumn{2}{c|}{Set14} & \multicolumn{2}{c|}{BSD100} & \multicolumn{2}{c}{Urban100} \\
                   &    &   & (M)   & PSNR   & SSIM    & PSNR   & SSIM    & PSNR   &  SSIM  & PSNR &  SSIM \\   
\hline
RSWS  & \multirow{2}*{\ding{55}} & \multirow{2}*{\ding{55}} & \multirow{2}*{0.29-0.33} 
&  30.14      & 0.8730      &  26.92       & 0.7685      &  26.53      & 0.7273      &  24.40       & 0.7419 \\
$N=30$ &&&        
& $\pm$0.1686 & $\pm$0.0016 &  $\pm$0.0929 & $\pm$0.0017 & $\pm$0.0379 & $\pm$0.0017 &  $\pm$0.0513 & $\pm$0.0025 \\
\hline
\multirow{2}*{\OurMethod}    & \multirow{2}*{\ding{55}} & \multirow{2}*{\ding{55}} & \multirow{2}*{0.28-0.29} 
&  30.76      & 0.8812      &  27.31       & 0.7761      &  26.92      & 0.7357      &  24.70       & 0.7517 \\&&&        
& $\pm$0.2083 & $\pm$0.0013 &  $\pm$0.1068 & $\pm$0.0013 & $\pm$0.0819 & $\pm$0.0020 &  $\pm$0.0885 & $\pm$0.0016 \\

\multirow{2}*{\OurMethod}    & \multirow{2}*{\ding{51}} & \multirow{2}*{\ding{55}} & \multirow{2}*{0.27-0.31} 
&  30.88      & 0.8825      &  27.44       & 0.7789      &  27.03      & 0.7372      &  24.87       & 0.7563 \\&&&        
& $\pm$0.1486 & $\pm$0.0012 &  $\pm$0.0222 & $\pm$0.0012 & $\pm$0.0814 & $\pm$0.0015 &  $\pm$0.0776 & $\pm$0.0025 \\

\multirow{2}*{\OurMethod}    & \multirow{2}*{\ding{55}} & \multirow{2}*{\ding{51}} & \multirow{2}*{0.29-0.32} 
&  31.66      & 0.8922      &  27.83       & 0.7876      &  27.32      & 0.7456      &  25.45       & 0.7760 \\&&&        
& $\pm$0.0115 & $\pm$0.0003 &  $\pm$0.0058 & $\pm$0.0002 & $\pm$0.0115 & $\pm$0.0001 &  $\pm$0.0208 & $\pm$0.0005 \\

\multirow{2}*{\OurMethod}    & \multirow{2}*{\ding{51}} & \multirow{2}*{\ding{51}} & \multirow{2}*{0.29-0.33} 
&  \bf{31.72} & \bf{0.8926} & \bf{27.85}   & \bf{0.7881} & \bf{27.35}  & \bf{0.7465} &  \bf{25.51}  & \bf{0.7779} \\&&&    
& $\pm$0.0404 & $\pm$0.0004 &  $\pm$0.0115 & $\pm$0.0004 & $\pm$0.0173 & $\pm$0.0007 &  $\pm$0.0289 & $\pm$0.0013 \\

\hline
\end{tabular}
}
\end{center}
\end{table*}

In this section, we focus on presenting
an 
ablation analysis on each component proposed in this paper. We first analyze the effects of
the 
layer-wise architecture sharing strategy and using residual learning.
Then we show the benefits of searching for the outer layer width.
Finally we verify %
whether 
the 
loss 
term
$\rm {l}_{ssim}$
indeed 
improves image restoration results. 

\subsubsection{Effects of layer-wise architecture sharing and residual learning}

To evaluate the effects of using the layer-wise architecture sharing strategy (LWAS) and using residual learning (ResL), we compare architectures founded in four different search settings. The first setting is %
using the basic search space designed in this paper. Based on the first setting, the second and third settings employ either layer-wise architecture sharing strategy or residual learning strategy. The last setting is using both strategies. The comparison results for image denoising and super-resolution are listed in Table \ref{tab: ablation_denoising} and Table \ref{tab: ablation_SR}, respectively. The corresponding visual results are shown in Figure \ref{fig:ab_lwas_resl_dn} and Figure \ref{fig:ab_lwas_resl_sr}.

From Table \ref{tab: ablation_denoising} and Table \ref{tab: ablation_SR}, we can see that both the layer-wise cell architecture sharing strategy and residual learning strategy improve the performance, and architectures using both strategies obtain the best performance. 

For images denosing, the layer-wise cell architecture sharing strategy and the residual learning strategy are equally important. For instance, with $\sigma=50$, the layer-wise architecture sharing strategy improves PSNR and SSIM by 0.05 dB and 0.0063, respectively. The residual learning strategy improves PNSR and SSIM by 0.06 dB and 0.0047, respectively. Compared with the layer-wise architecture sharing, the residual learning strategy improves more in PSNR and less in SSIM. Using both strategies improves PNSR and SSIM from 26.77 dB and 0.7584 to 26.84 dB and 0.7654, respectively. From Figure \ref{fig:ab_lwas_resl_dn}, we can see that denoising results of the last setting shows more details. 

For image super-resolution, from the first setting %
to the second setting, 
PSNR and SSIM show slight improvement. 
For the third  setting, 
PSNR and SSIM are significantly improved by using residual learning. The fourth setting %
again 
achieves the best performance. Taking the results on BSD100 as an example, PSNR and SSIM are improved to 27.08 dB and 0.7375 by using the layer-wise cell architecture sharing, 27.31 dB and 0.7455 by using the residual learning. Using both layer-wise cell architecture sharing, and residual learning achieves the best performance with PSNR $= 27.35$ dB and SSIM $=0.7467$. According to results shown in Figure \ref{fig:ab_lwas_resl_sr}, the produced results of the using both strategies are %
closest to the ground truths.

\subsubsection{Benefits of searching for the outer layer width}

\begin{figure}[t]
\begin{center}
\includegraphics[width=2.8 in]{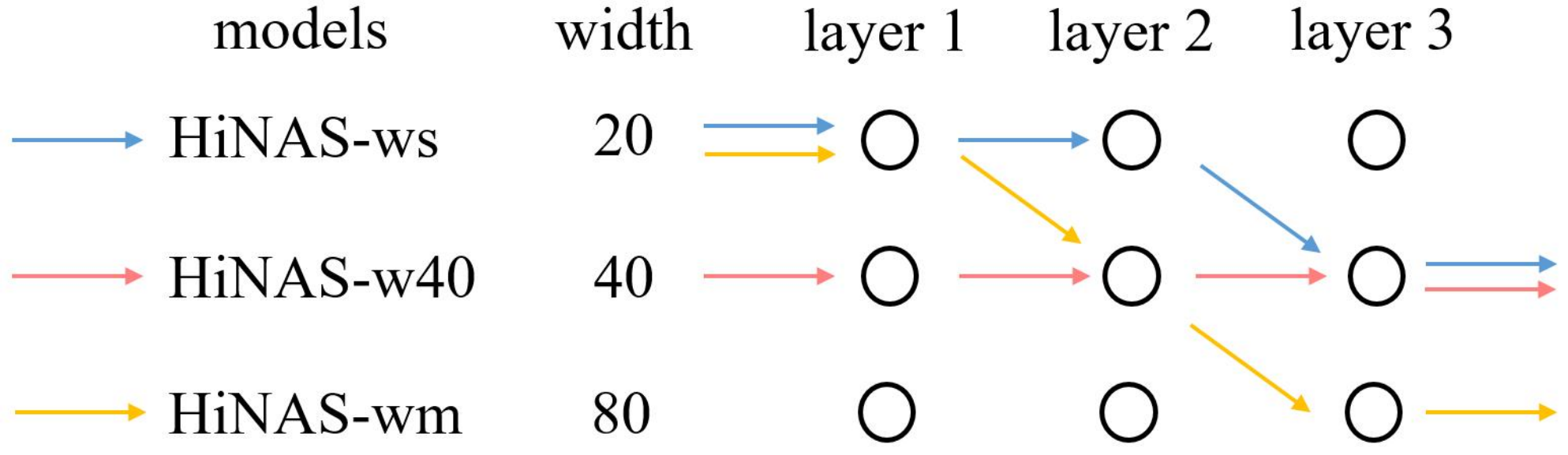}
\end{center}
\caption{Comparisons of different search settings.}
\label{fig:comparison_width_search}
\end{figure}

\begin{table}[t]
\caption{Comparisons of different search settings.}
\label{Table: comparisons of different search settings}
\footnotesize
\renewcommand\arraystretch{1.0}
\begin{center}
{
\begin{tabular}{ l  | ccc}

\hline
Models  &  \# parameters (M) & PSNR  & SSIM \\
\hline
\OurMethod-ws      & 0.34  & 29.25  & 0.8420 \\
\OurMethod-w40     & 0.57  & 29.14  & 0.8409 \\ 
\OurMethod-wm      & 0.75  & 29.19  & 0.8413 \\
\hline
\end{tabular}
}
\end{center}

\end{table}

In this section, to evaluate the benefits of searching outer layer width, we apply our \OurMethod on BSD500 with three different search settings, which are denoted as \OurMethod-ws, \OurMethod-w40, \OurMethod-wm. For \OurMethod-ws, both the inner cell architectures and out layer width are found by our \OurMethod algorithm. For the two latter  settings, only the inner cell architectures are found by our algorithm and the outer layer widths are set manually. The width of each cell are set to 40 for \OurMethod-w40. In \OurMethod-wm, we set the basic width of the first cell to 20, then double the basic width cell by cell. The three settings are shown in Figure~\ref{fig:comparison_width_search}. The comparison results for denoising on BSD500 of $\sigma=30$  are listed in Table~\ref{Table: comparisons of different search settings}. 

As shown in Table~\ref{Table: comparisons of different search settings}, from \OurMethod-ws to \OurMethod-w40, PSNR and SSIM decrease from 29.25 dB and 0.8420 to 29.14 dB and 0.8409, respectively. In addition, the number of parameters increases from 0.34M to 0.57M. From \OurMethod-w40 to \OurMethod-wm, PSNR and SSIM show slight improvement: 0.05 dB for PSNR and 0.0004 for SSIM.  Meanwhile, the corresponding number of parameters is increased to 0.75M.

With searching for the outer layer width, \OurMethod-ws achieves the best performance, while having the least parameters.

\subsubsection{Benefits of using the ${\rm {l}_{ssim}}$ loss}

\begin{figure*}[h]
\begin{center}
\includegraphics[width=.7158\textwidth]{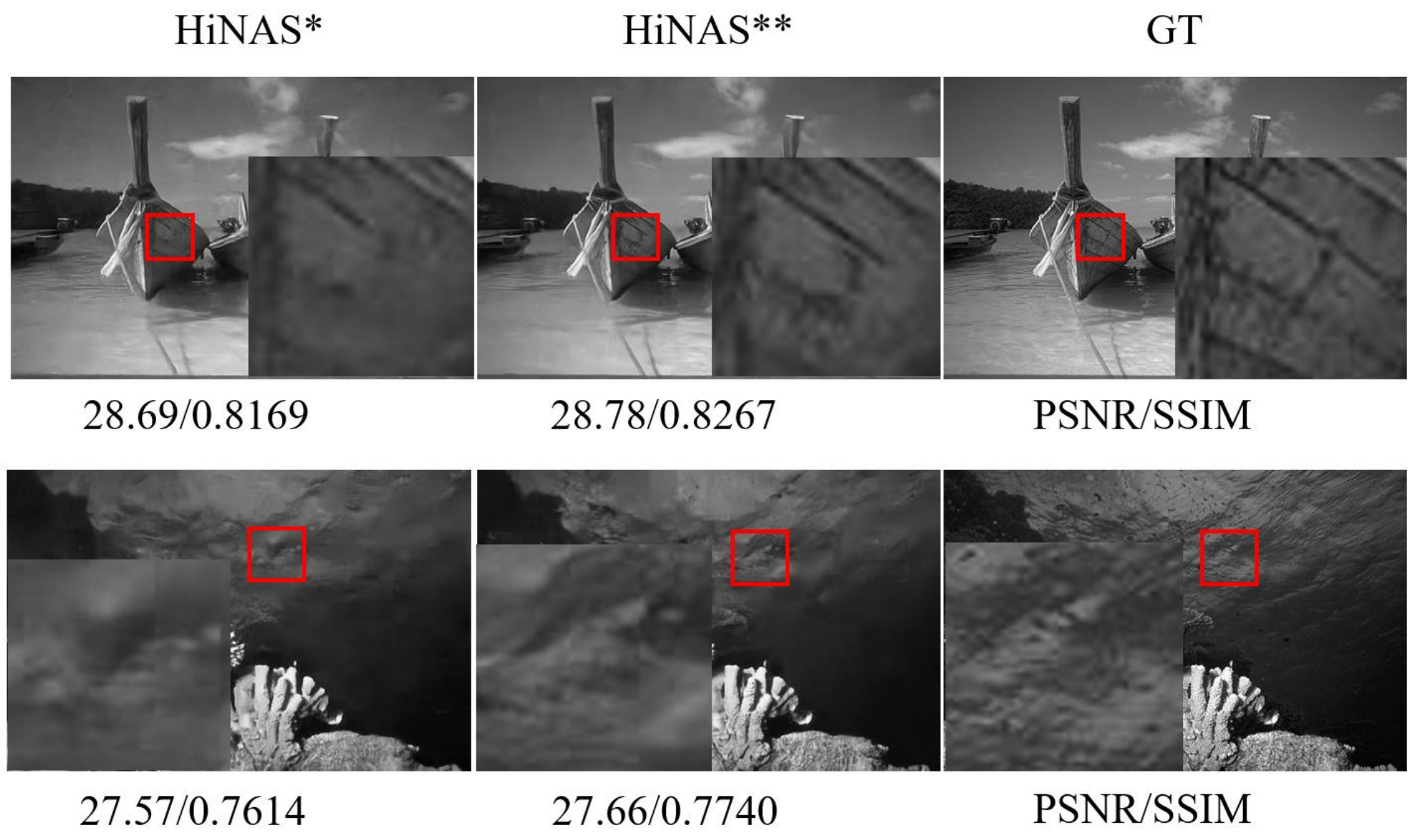}
\end{center}
\caption{Visual results of image denoising. The first row shows the denoising results for image `81095' from BSD200 when $\sigma=50$. The second row is the denoising results for the image `101027' from BSD200 when $\sigma=50$. Best viewed on screen.
}
\label{fig:ab_loss_dn}
\end{figure*}

\begin{figure*}[t]
\begin{center}
\includegraphics[width=.8\textwidth]{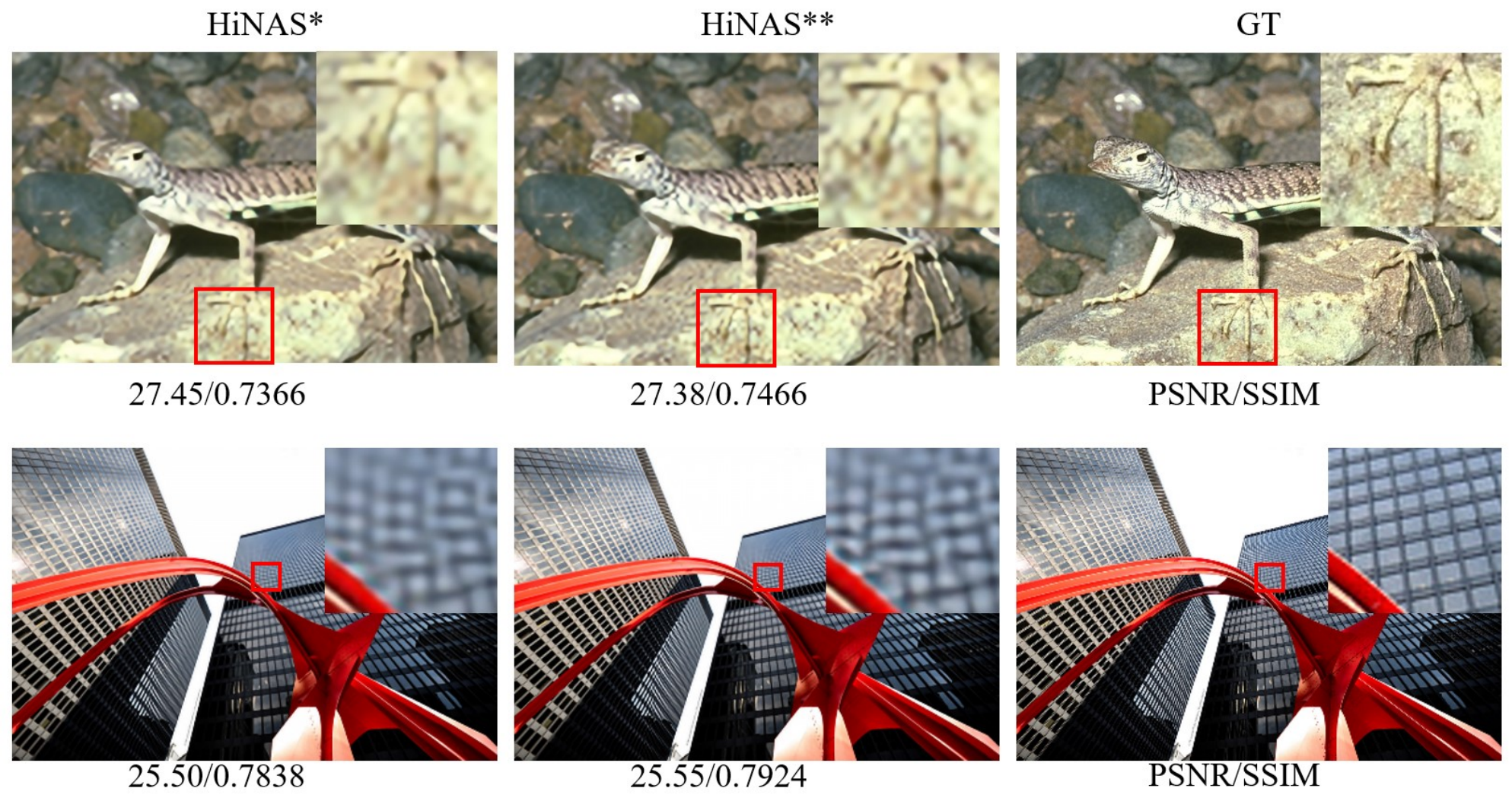}
\end{center}
\caption{Visual results of  4$\times$ image super-resolution. The two rows are the results for image `87046' from BSD100 and image `062' from Urban100.}
\label{fig:ab_loss_sr}
\end{figure*}

\newcolumntype{C}[1]{>{\centering\let\newline\\\arraybackslash\hspace{-11pt}}m{#1}}
\begin{table}[t!]
\caption{Benefits of using $\rm {l}_{ssim}$ loss item for image denoising. \OurMethod\!\!$ ^ *$ is trained with single loss MSE and \OurMethod\!\!$ ^{**} $ is trained with the combination loss MSE and ${\rm {l}_{ssim}}$. The last row is the comparison results. Blue and red denote the performance increase and decrease over \OurMethod\!\!$ ^ *$.}
\label{Table: loss_dn}
\footnotesize
\renewcommand\arraystretch{1.0}
\begin{center}
{
\begin{tabular}{l C{0.7cm}C{0.65cm}C{0.65cm}C{0.65cm}C{0.65cm}C{0.65cm}}
\hline
 \multirow{2}*{Method}  &   \multicolumn{2}{c}{$\sigma=30$} & \multicolumn{2}{c}{$\sigma=50$} &  \multicolumn{2}{c}{$\sigma=70$}  \\
         &  PSNR  & SSIM   &  PSNR  & SSIM   &  PSNR  & SSIM \\
\hline
\OurMethod\!\!$^*$ & 29.03  & 0.8254  & 26.77  & 0.7498   & 25.42  & 0.6962  \\ 
\OurMethod\!\!$^{**}$ & 29.14  & 0.8403  & 26.77  & 0.7635  & 25.48  & 0.7129 \\ 
Comparison  & \textcolor{blue}{0.11} & \textcolor{blue}{0.0149}  &\textcolor{blue}{0.009} & \textcolor{blue}{0.0137} & \textcolor{blue}{0.06} & \textcolor{blue}{0.0167}\\

\hline
\end{tabular}
}
\end{center}

\end{table}

\newcolumntype{C}[1]{>{\centering\let\newline\\\arraybackslash\hspace{-11pt}}m{#1}}
\begin{table}[b]
\caption{Benefits of using $\rm l_{\rm ssim}$ loss item for $\times$4 image super-resolution. }
\label{Table: loss_sr}
\footnotesize
\renewcommand\arraystretch{1.0}
\begin{center}
{
\begin{tabular}{l C{0.7cm}C{0.65cm}C{0.65cm}C{0.65cm}C{0.65cm}C{0.65cm}}
\hline
 \multirow{2}*{Methods}  &   \multicolumn{2}{c}{Set5} & \multicolumn{2}{c}{BSD100} &  \multicolumn{2}{c}{Urban100}  \\
         &  PSNR  & SSIM   &  PSNR  & SSIM   &  PSNR  & SSIM \\
\hline
\OurMethod\!\!$^*$ & 31.72  & 0.8904  & 27.38  & 0.7367   & 25.47  & 0.7691  \\ 
\OurMethod\!\!$^{**}$ & 31.74  & 0.8928  & 27.35  & 0.7467   & 25.52  & 0.7785 \\ 
Comparison            & \textcolor{blue}{0.02}   & \textcolor{blue}{0.0024}  & \textcolor{red}{0.03}   &  \textcolor{blue}{0.01}    & \textcolor{blue}{0.05}   & \textcolor{blue}{0.0094} \\

\hline
\end{tabular}
}
\end{center}
\end{table}

\begin{figure*}[t]
\begin{center}
\includegraphics[width=\textwidth]{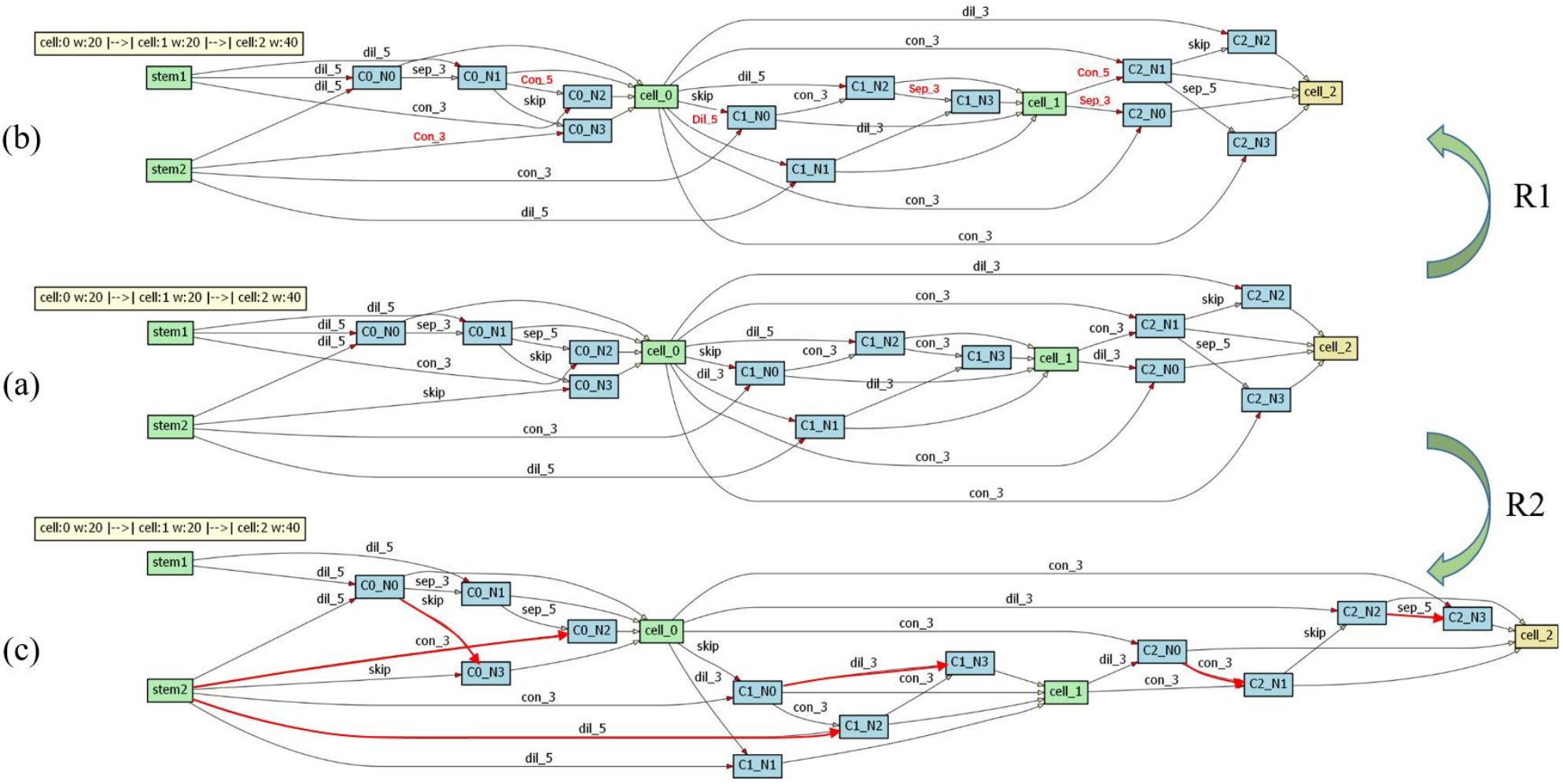}
\end{center}
\caption{Architecture analysis for image denoising. `Conv', `sep' and `dil' denote conventional, separable and dilated convolutions. `Skip' is skip connection. (a) The architecture found by \OurMethod (b) modified cells, R1; (c) modified cells, R2. The operations and paths marked in red denote modification parts}
\label{fig:architecture_analysis_dn}
\end{figure*}

\begin{figure*}[t]
\begin{center}
\includegraphics[width=.82706\textwidth]{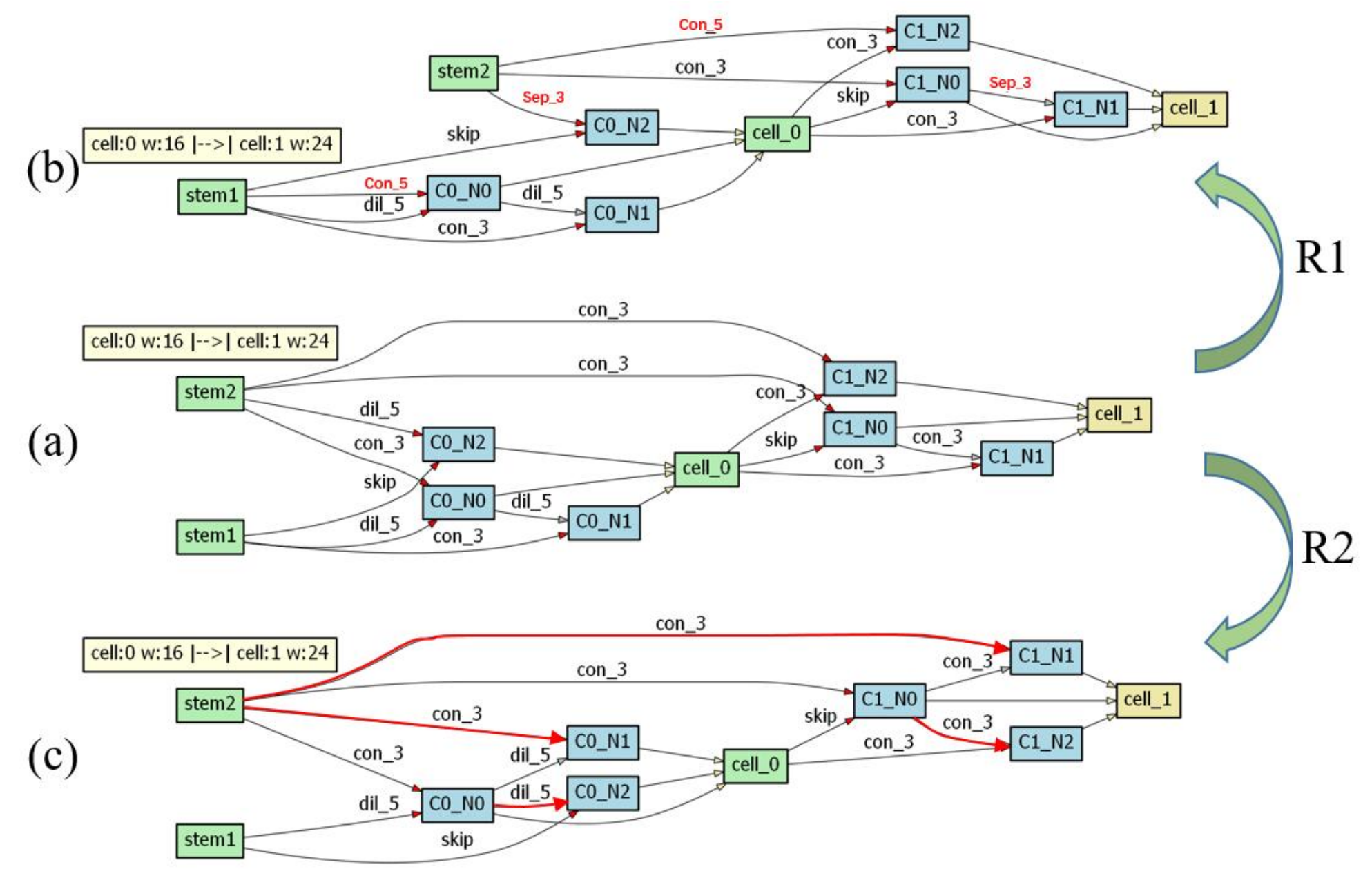}
\end{center}
\caption{Architecture analysis for image super-resolution. (a) The architecture found by \OurMethod (b) modified cells, $\rm R1$; (c) modified cells, $\rm R2$. The operations and paths marked in red denote modification parts}
\label{fig:architecture_analysis_sr}
\end{figure*}

Here, we analyze how our designed loss item ${\rm {l}_{ssim}}$ improves image restoration results. We implement two baselines: 1) \OurMethod\!\!$^*$ trained with single MSE loss; and 2) \OurMethod\!\!$^{**}$  trained with the combination MSE loss and ${\rm {l}_{ssim}}$. 

The experimental results of image denoising on B\-S\-D\-5\-0\-0 dataset with $\sigma=30$ are listed in Table \ref{Table: loss_dn} and shown in Figure \ref{fig:ab_loss_dn}, from which we can see that \OurMethod\!\!$^{**}$ achieves better results. Comparing the denosing results in Figure \ref{fig:ab_loss_dn}, we can find that the denoising images of \OurMethod\!\!$^{**}$ reverse more rich texture information as compared with that of \OurMethod\!\!$^{*}$. Note that even trained with single MSE loss, \OurMethod\!\!$^{*}$ still outperforms the competitive models, such as N3Net, NLRN, etc.

The experimental results of $\times$4 image super-resolution are reported in Table \ref{Table: loss_sr}. Figure \ref{fig:ab_loss_sr} shows the corresponding visual results. 

For image denoising, the loss %
term 
$\rm {l}_{ssim}$ benefits both PSNR and SSIM metrics. However, for image super-resolution, the loss %
term 
$\rm {l}_{ssim}$ does not always improve  both PSNR and SSIM. On BSD100, using the loss %
term 
$\rm {l}_{ssim}$ improves SSIM by 0.01, but decreases PSNR by 0.03dB. On Set5 and Urban100, \OurMethod\!\!$^{**}$ trained with the combination loss shows even better results over \OurMethod\!\!$^{*}$. On the other dataset Set14 which is not listed in here, \OurMethod\!\!$^{**}$ also shows better performance than \OurMethod\!\!$^{*}$. We conjecture that this is caused by the property of ${l}_{ssim}$, which is designed based on SSIM evaluation metric. Using $\rm {l}_{ssim}$ %
would 
improve SSIM directly and affect PSNR indirectly. The changes of SSIM and PSNR are not always consistent. Figure \ref{fig:ab_loss_sr} shows two examples, and the images in the  first row are come from BSD100. Comparing the middle image with the image on the left side in the first row, we can see that, \OurMethod\!\!$^{**}$ has higher SSIM and lower PSNR compared with \OurMethod\!\!$^{*}$. However, from visual results, we can not see any difference. In the second row, the middle image shows cleaner grid structure than the image on the left side. In summary, for image super-resolution, $\rm {l}_{ssim}$ still %
improves 
the performance. 

\begin{table}[t]
\caption{Architecture analysis for image denoising.}
\label{Table: architecture_analysis_dn}
\footnotesize
\renewcommand\arraystretch{1.0}
\begin{center}
{
\begin{tabular}{l|ccc}
\hline
Methods &  \OurMethod  & \OurMethod, $\rm R1$ & \OurMethod, $\rm R2$ \\
\hline

PSNR    &  \textbf{29.25}   & 29.21  &  29.12  \\
SSIM    &  \textbf{0.8420}  & 0.8418 &  0.8304 \\
\hline
\end{tabular}
}
\end{center}
\end{table}

\newcolumntype{C}[1]{>{\centering\let\newline\\\arraybackslash\hspace{-11pt}}m{#1}}
\begin{table}[h]
\caption{Architecture analysis for super-resolution.}
\label{Table: architecture_analysis_sr}
\footnotesize
\renewcommand\arraystretch{1.0}
\begin{center}
{
\begin{tabular}{l C{0.7cm}C{0.65cm}C{0.65cm}C{0.65cm}C{0.65cm}C{0.65cm}}
\hline
 \multirow{2}*{Methods}  &   \multicolumn{2}{c}{Set5} & \multicolumn{2}{c}{BSD100} &  \multicolumn{2}{c}{Urban100}  \\
         &  PSNR  & SSIM   &  PSNR  & SSIM   &  PSNR  & SSIM \\
\hline
\OurMethod & {\bf 31.74}  & {\bf 0.8928}  & {\bf 27.35}  & {\bf 0.7467}   & {\bf 25.52}  & {\bf 0.7785} \\  
\OurMethod, R1 & 31.70  & 0.8926  & 27.34  & 0.7458   & 25.52  & 0.7782 \\
\OurMethod, R2 & 31.58  & 0.8821  & 27.13  & 0.7312   & 25.13  & 0.7585 \\

\hline
\end{tabular}
}
\end{center}
\end{table}

\begin{table*}[h]
\footnotesize
\renewcommand\arraystretch{1.2}
\renewcommand\tabcolsep{4.0 pt}
\begin{center}
{
\begin{tabular}{l|c|c|cc|cc|cc|cc|c}
\hline
\multirow{2}*{Methods} & cell & \# param. &  \multicolumn{2}{c|}{$\sigma$ = 30} & \multicolumn{2}{c|}{$\sigma$ = 50} & \multicolumn{2}{c|}{$\sigma$ = 70}  & \multicolumn{2}{c|}{search cost} & search\\
   & sharing & (M)  &  PSNR  & SSIM  & PSNR  & SSIM & PSNR & SSIM & GPU  & hours & method  \\
\hline
E-CAE & - & 1.10  &  28.23 & 0.8047 & 26.17 & 0.7255 & 24.83  & 0.6636  & 4 V100 & 44.0  & EA \\
\hline
\multirow{2}*{\OurMethod} & \multirow{2}*{\ding{55}}  & \multirow{2}*{0.23-0.34}  
&  29.21      & 0.8403      & 26.75       & 0.7620      & 25.38       & 0.7049 & 
\multirow{2}*{1 1080Ti}  & \multirow{2}*{3.6} & \multirow{2}*{gradient} \\ & & 
& $\pm$0.0336 & $\pm$0.0016 & $\pm$0.0603 & $\pm$0.0034 & $\pm$0.0213 & $\pm$0.0011  &  & \\
\multirow{2}*{\OurMethod} & \multirow{2}*{\ding{51}}  & \multirow{2}*{0.22-0.34} 
& \bf{29.23}  & \bf{0.8411} & \bf{26.81}  & \bf{0.7649} & \bf{25.43}  & \bf{0.7069} & 
\multirow{2}*{1 1080Ti}  & \multirow{2}*{1.0} & \multirow{2}*{gradient} \\ & &
& $\pm$0.0346 & $\pm$0.0015 & $\pm$0.0700 & $\pm$0.0035 & $\pm$0.0265 & $\pm$0.0006  &  & \\
\hline
\end{tabular}
}
\end{center}
\caption{Comparisons with NAS methods of image denoising. For E-CAE, the search and training time costs computed on V100 GPUs are provided by authors.}
\vspace{-0.4 cm}
\label{Table: comparison_with_NASs_dn}
\end{table*}

\begin{table*}[h]
\footnotesize
\renewcommand\arraystretch{1.2}
\renewcommand\tabcolsep{3.04 pt}
\begin{center}
{
\begin{tabular}{l|c|c|cc|cc|cc|cc|cc|c}
\hline
\multirow{2}*{Methods} & cell & \# param. &  \multicolumn{2}{c|}{Set5} & \multicolumn{2}{c|}{BSD100} & \multicolumn{2}{c|}{Urban100}  & \multicolumn{2}{c|}{search cost} & search\\
    & sharing    & (M)  &  PSNR  & SSIM  & PSNR  & SSIM & PSNR & SSIM & GPU  & hours & method  \\
\hline
MoreMNAS & - & 1.04  &  37.63 & 0.9584 & 31.95 & 0.8961 & 31.24  & 0.9187  & 8 V100 & 168 & RL + EA \\
FALSR  & -  & 0.41  &  37.66 & 0.9586 & 31.96  & 0.8965 & 31.24 & 0.9187 & 8 V100 & 72  & RL + EA \\
\hline
\multirow{2}*{\OurMethod} & \multirow{2}*{\ding{55}} & \multirow{2}*{0.24-0.28} 
& \bf{37.66}  & 0.9606      & \bf{31.98}   & \bf{0.9036} &  31.27       &  0.9233   & 
\multirow{2}*{1 1080Ti} & \multirow{2}*{12.6} & \multirow{2}*{gradient} \\ & & 
& $\pm$0.0299 & $\pm$0.0002 & $\pm$0.0200 & $\pm$0.0001 &  $\pm$0.0492 & $\pm$0.0011 & &\\
\multirow{2}*{\OurMethod} & \multirow{2}*{\ding{51}} & \multirow{2}*{0.23-0.28} 
& 37.65       & \bf{0.9608} & 31.97       & 0.9036      &  \bf{31.29}  & \bf{0.9236}   & 
\multirow{2}*{1 1080Ti}  & \multirow{2}*{3.5} & \multirow{2}*{gradient} \\ & &
& $\pm$0.0298 & $\pm$0.0001 & $\pm$0.0222 & $\pm$0.0003 & $\pm$0.0200  &  0.0003   & & \\
\hline
\end{tabular}
}
\end{center}
\caption{Comparisons with NAS methods of image super-resolution. }
\vspace{-0.4 cm}
\label{Table: comparison_with_NASs_sr}
\end{table*}

\subsubsection{Architecture analysis}

The denoising and suepr-resolution architectures found by \OurMethod are shown in Figure \ref{fig:architecture_analysis_dn}(a) and Figure \ref{fig:architecture_analysis_sr}(a), from which we can see that:

\begin{enumerate}
\itemsep 0pt
    \item In the both denoising and suepr-resolution networks found by our \OurMethod, the width of the last layer has the maximum number of channels. This is consistent  with previous manually designed networks. 
    
    \item Instead of connecting all the nodes with a certain convolution, \OurMethod connects different nodes with different types of operators. We believe that these results prove that \OurMethod is able to select proper operators.
    
\end{enumerate}

From Figure \ref{fig:architecture_analysis_dn}(a) and Figure \ref{fig:architecture_analysis_sr}(a), we also find that the networks found  by \OurMethod consist of many fragmented branches, which might be the main reason why the designed networks have better performance than previous denoising models. As explained in \citep{ma2018shufflenet}, the fragmentation structure is beneficial for accuracy. Here we verify if \OurMethod improves the accuracy by designing a proper architecture or by simply integrating various branch structures and convolution operations.

We modify the architecture found by our \OurMethod in two different ways, and then compare the modified architectures with unmodified architectures.

The first modification is replacing conventional convolutions in the searched architectures with other convolutions to verify is the operation that \OurMethod selected for each path is irreplaceable, as shown in Figure~\ref{fig:architecture_analysis_dn}(b) and Figure~\ref{fig:architecture_analysis_sr}(b).

The other modification is to change the topological structure inside each cell, as shown in Figure~\ref{fig:architecture_analysis_dn}(c) Figure~\ref{fig:architecture_analysis_sr}(c), which is aiming to verify if the connection relationship built by our \OurMethod is indeed appropriate.

Following the two proposed modifications, we modify different operations and connections in different nodes. Modified architectures achieve lower performance. However, limited by space, we only show two examples here. The modification parts are marked in red. The comparison results are listed in Tables~\ref{Table: architecture_analysis_dn} and~\ref{Table: architecture_analysis_sr}, where the two mentioned modification operations, are denoted as $R1$ and $R2$. From Tables~\ref{Table: architecture_analysis_dn} and~\ref{Table: architecture_analysis_sr}, we can see that both modifications %
degrade 
the accuracy for image denoising and super-resolution. Replacing convolution operations slightly reduces PSNR and SSIM. Compared with replacing convolution operations, changing topological structures is more influential on the performance.  

From the comparison results, we can draw a conclusion:  \OurMethod does find a proper structure and selects proper convolution operations, instead of simply integrating a complex network with various operations. \textit{The fact that a slight perturbation to the found architecture deteriorates the accuracy indicates that the found architecture is indeed a local optimum in the architecture search space. }

\begin{figure*}[t]
\begin{center}
\includegraphics[width=6.6 in]{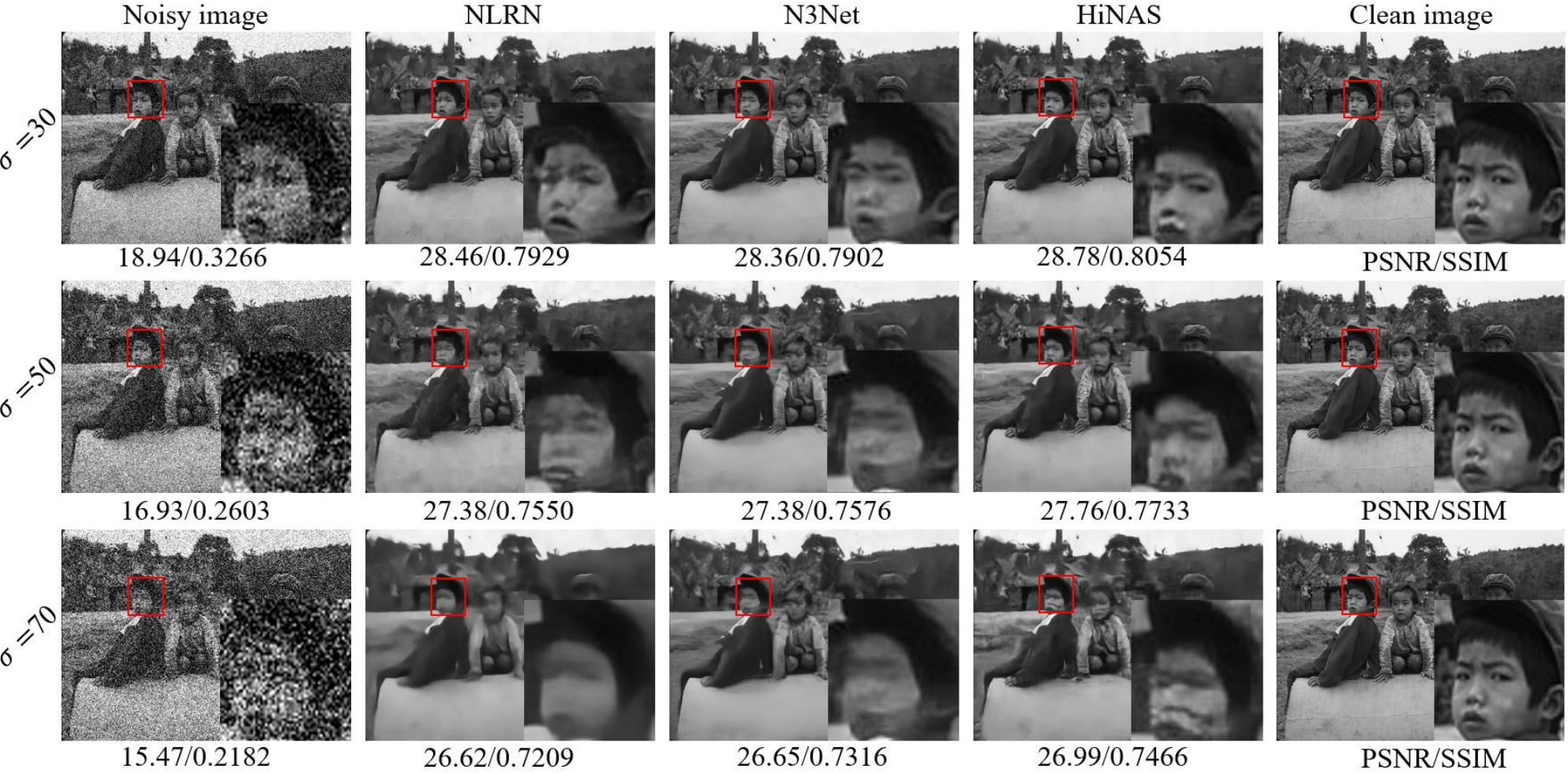}
\end{center}
\caption{ Visual denoising results of NLRN, N3Net and \OurMethod. The first row is the denoising results of image `15011' from BSD 200 when $\sigma=30$. The second row and the third row are the denoising results of $\sigma=50$ and $\sigma=70$.}
\label{fig:comparison_sota_dn}
\end{figure*}

\begin{table*}[h]
\caption{Denoising experiments. Comparisons with state-of-the-arts on the BSD500 dataset. We show our results in the last row. Time cost means GPU-seconds for inference on the 200 images from the test set of BSD500 using one single GTX 1080Ti graphic card. The 
best and second-best results are marked in red and blue. Here, we compare the proposed method with 9 other methods, including BM3D~\cite{dabov2007image}, WNNM~\cite{gu2014weighted}, RED~\cite{mao2016image}, MemNet~\cite{tai2017memnet}, NLRN~\cite{liu2018non}, E-CAE~\cite{suganuma2018exploiting}, DBSN~\cite{wu2020unpaired}, DuRN-P~\cite{liu2019dual} and N3Net~\cite{plotz2018neural}.
}
\label{Table: BSD500}
\footnotesize
\renewcommand\arraystretch{1.0}
\begin{center}
{
\begin{tabular}{l|c|c|cc|cc|cc}
\hline
\multirow{2}*{Methods}  & \multirow{2}*{\# parameters (M)}  & \multirow{2}*{time cost (s)} & \multicolumn{2}{c|}{$\sigma=30$} & \multicolumn{2}{c|}{$\sigma=50$} & \multicolumn{2}{c}{$\sigma=70$} \\
                             &      &     & PSNR   & SSIM    & PSNR   & SSIM    & PSNR   &  SSIM    \\   
\hline
BM3D &   -  &  -  & 27.31  & 0.7755  & 25.06  & 0.6831  & 23.82  &  0.6240  \\
WNNM &   -  &  -  & 27.48  & 0.7807	 & 25.26  &	0.6928	& 23.95  &  0.3460  \\
RED  &   -  &  -  & 27.95  & 0.8056  & 25.75  &	0.7167	& 24.37  &  0.6551  \\
MemNet & \textcolor{blue}{0.68} &  \textcolor{blue}{32.67}  & 28.04  & 0.8053	 & 25.86  &	0.7202	& 24.53  &  0.6608  \\
NLRN    & 1.03 &  6940.67  & 28.15  & \textcolor{red}{0.8423} & 25.93  & 0.7214	& 24.58  &  0.6614  \\
E-CAE  & 1.10 & - & 28.23 & 0.8047  & 26.17  & 0.7255  & 24.83  & 0.6636 \\
DBSN  & 6.61 &  127.56   & -  & -   & 26.18  & 0.7257  &  -  &  -  \\
DuRN-P & 0.82 &  -  & 28.50  & 0.8156  & 26.36  & 0.7350  & 25.05  &  0.6755\\
N3Net & 0.71 &  80.74 & \textcolor{blue}{28.66}  & 0.8220  & \textcolor{blue}{26.50}  & \textcolor{blue}{0.7490}	& \textcolor{blue}{25.18}  &  \textcolor{blue}{0.6960}  \\
\OurMethod & \textcolor{red}{0.22-0.34} & \textcolor{red}{16.21} & \textcolor{red}{29.23} & \textcolor{blue}{0.8411} & \textcolor{red}{26.81}  & \textcolor{red}{0.7649} & \textcolor{red}{25.43} & \textcolor{red}{0.7069} \\
\hline
\end{tabular}
}
\end{center}

\end{table*}

\begin{figure*}[t]
\begin{center}
\includegraphics[width=6.6 in]{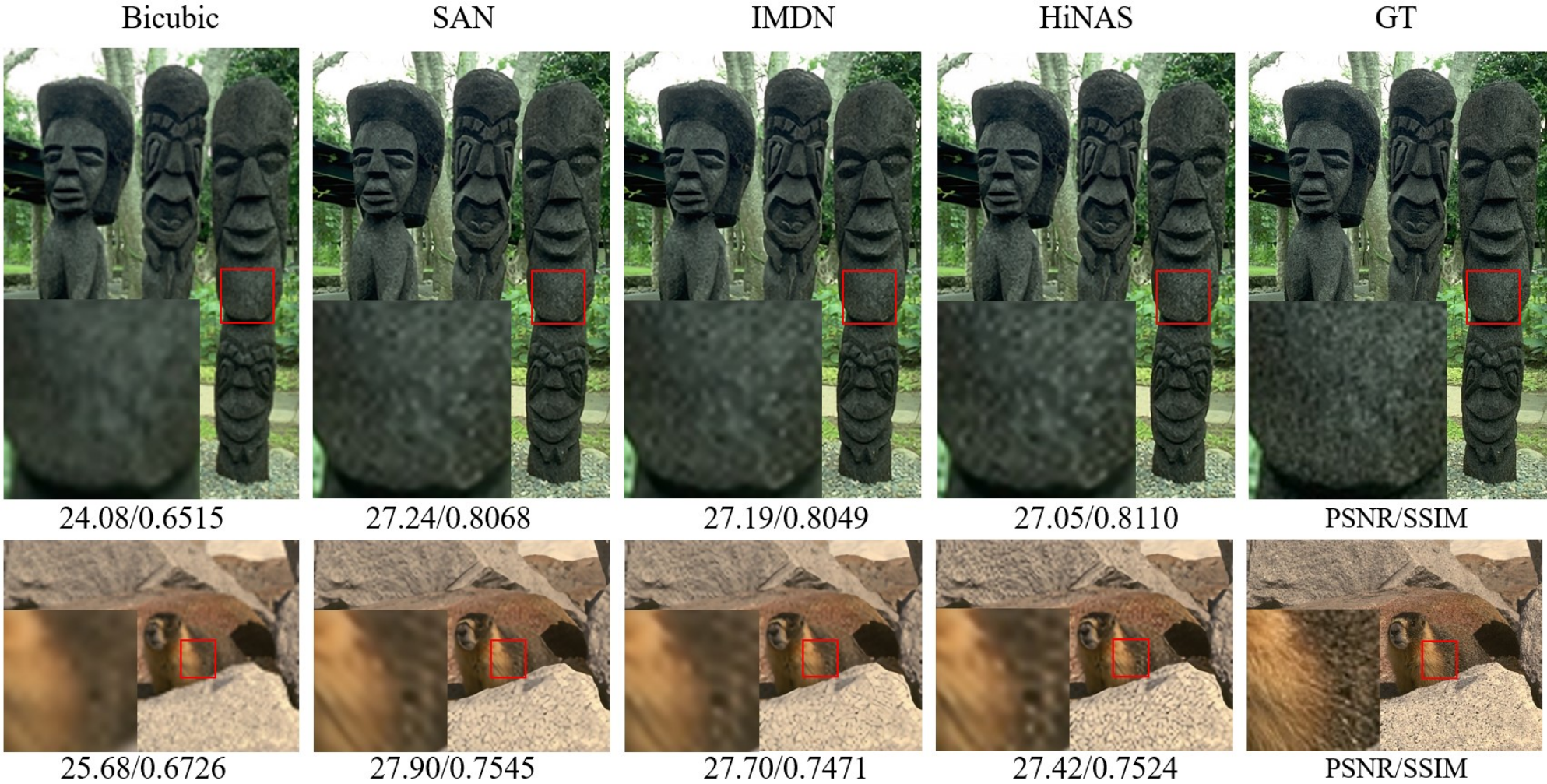}
\end{center}
\caption{Visual super-resolution results of Bicubic, SAN, IMDN and \OurMethod. First row: the $\times$2 super-resolution results for image `101085' from BSD100. Second row: the $\times$4 super-resolution results for image '41069' from BSD100. }
\label{fig:comparison_sota_sr}
\end{figure*}

\begin{table*}[h]
\caption{Super-resolution experiments with the scale factor being 3.  We show our results in the last row. Time cost means GPU-seconds for inference on the 100 images from BSD100 using one single GTX 1080 Ti graphic card. The best and second-best results are marked in red and blue. Here, the proposed method is compared with other 12 approaches, including EDSR~\cite{lim2017enhanced}, SRMD~\cite{zhang2018learning}, RDN~\cite{zhang2018residual}, RCAN~\cite{zhang2018image}, SAN~\cite{dai2019second}, RFANet~\cite{liu2020residual}, FSRCNN(KD)~\citep{lee2020learning}, VDSR~\citep{kim2016accurate}, LapSRN~\citep{lai2017deep}, MemNet~\citep{tai2017memnet}, IDN~\citep{hui2018fast} and IMDN~\citep{hui2019lightweight}.
}
\label{tab: SR_experiments}
\footnotesize
\renewcommand\arraystretch{1.0}
\begin{center}
{
\begin{tabular}{l|c|c|cc|cc|cc|cc}
\hline
\multirow{2}*{Method} & \multirow{2}*{Param. (M)} & \multirow{2}*{time cost (s)} & \multicolumn{2}{c|}{Set5} & \multicolumn{2}{c|}{Set14} & \multicolumn{2}{c|}{BSD100} & \multicolumn{2}{c}{Urban100} \\
   &     &   &  PSNR  & SSIM &  PSNR  & SSIM &  PSNR  & SSIM &  PSNR  & SSIM \\
\hline
EDSR     & 43.00 & - & 34.65 & 0.9280 & 30.52 & 0.8462 & 29.25 & 0.8093 & 28.80 & 0.8653  \\
SRMD        & 1.50 & - & 34.12 & 0.9254 & 30.04 & 0.8382 & 28.97 & 0.8025 & 27.57 & 0.8398 \\
RDN         & 22.30 & 11.89 & 34.71 & 0.9296 & 30.57 & 0.8468 & 29.26 & 0.8093 & 28.80 & 0.8653  \\
RCAN       & 16.00 & 14.60 & 34.74 & 0.9299 & 30.64 & 0.8481 & 29.32 & 0.8111 & 29.08 & 0.8702  \\
SAN         & 15.70 & 34.59 & 34.75 & 0.9300 & 30.59 & 0.8476 & 29.33 & 0.8112 & 28.93 & 0.8671  \\
RFANet      & 11.00 & - & 34.79 & 0.9300 & 30.67 & 0.8487 & 29.34 & 0.8115 & 29.1 & 0.872  \\
\hline
bicubic    & - & - & 30.39 & 0.8682 & 27.55 & 0.7742 & 27.21 & 0.7385 & 24.46 & 0.7349  \\
FSRCNN(KD) & \textcolor{red}{0.013} & \textcolor{red}{0.58} & 33.31 & 0.9179 & 29.57 & 0.8276 & 28.61 & 0.7919 & 26.67 & 0.8153  \\
VDSR       & 0.67 & - & 33.67 & 0.9210 & 29.78 & 0.8320 & 28.83 & 0.7990 & 27.14 & 0.8290  \\
LapSRN     & 0.50 & 2.94 & 33.82 & 0.9227 & 29.87 & 0.8320 & 28.82 & 0.7980 & 27.07 & 0.8280  \\
MemNet     & 0.68 & 7.37 & 34.09 & 0.9248 & \textcolor{blue}{30.01} & 0.8350 & \textcolor{blue}{28.96} & 0.8001 & \textcolor{blue}{27.56} & 0.8376  \\
IDN        & 0.55 & 2.90 & \textcolor{blue}{34.11} & \textcolor{blue}{0.9253} & 29.99 & 0.8354 & 28.95 & 0.8013 & 27.42 & 0.8359  \\
IMDN       & 0.70 & 1.91 & \textcolor{red}{34.36} & \textcolor{red}{0.9270} & \textcolor{red}{30.32} & \textcolor{blue}{0.8417} & \textcolor{red}{29.09} & \textcolor{blue}{0.8046} & \textcolor{red}{28.17} & \textcolor{red}{0.8519}  \\
\OurMethod & \textcolor{blue}{0.25-0.30} & \textcolor{blue}{1.59} & 33.95 &	0.9265 &	29.59 &	\textcolor{red}{0.8465} & 28.84 & \textcolor{red}{0.8135} & 27.44	& \textcolor{blue}{0.8448} \\
\hline

\end{tabular}
}
\end{center}
\end{table*}

\subsection{Comparisons with Other NAS Methods}
\label{Sec: comparisons with NAS methods}

Inspired by recent advances in NAS, four NAS methods have been proposed for low-level image restoration tasks \citep{suganuma2018exploiting, chu2019fast, liu2019deep}. E-CAE \citep{suganuma2018exploiting} is proposed for image inpainting and denoising. MoreMNAS \citep{chu2019multi} and FALSR \citep{chu2019fast} are proposed for super resolution. EvoNet \citep{liu2019deep} searches for networks for medical image denoising. In this section, we first compare our \OurMethod with E-CAE for searching for architectures for image denoising on BSD500. Then we compare our \OurMethod with MoreMNAS and FALSR for searching super-resolution networks on DIV2K. Tables~\ref{Table: comparison_with_NASs_dn} and~\ref{Table: comparison_with_NASs_sr} show the details.

Compared with previous NAS methods for low-level image processing tasks, our \OurMethod is much faster in searching. For instance, by using 8 Tesla V100 GPUs, FALSR takes about 72 hours to find the best super-resolution architecture on DIV2K and MoreMNAS takes 168 hours. In contrast to them, \OurMethod  only  takes about 3.5 hours with one GTX 1080Ti GPU. The fast search speed of our \OurMethod benefits from the following three advantages. 
\begin{enumerate}

    \item \OurMethod uses a gradient based search strategy. E-CAE, MoreMNAS and FALSR need to train many children networks (genes) to update their populations or controllers. For instance, FALSR trained about 10k models during its searching process. In sharp contrast, our \OurMethod only needs to train one supernet in the search stage. 
    
    \item In searching for the outer layer width, we share cells across different feature levels, saving  memory consumption in the supernet. As a result, we can use larger batch sizes for training the supernet, which further speeds up search. Comparing the last two rows of Tables~\ref{Table: comparison_with_NASs_dn} and~\ref{Table: comparison_with_NASs_sr}, we can see that the proposed cell sharing strategy significantly accelerates the search speed without any negative influence to performance. 
    
\end{enumerate}

\subsection{Comparisons with State-of-the-art Methods}

\begin{figure*}[h]
\begin{center}
\includegraphics[width=6.6 in]{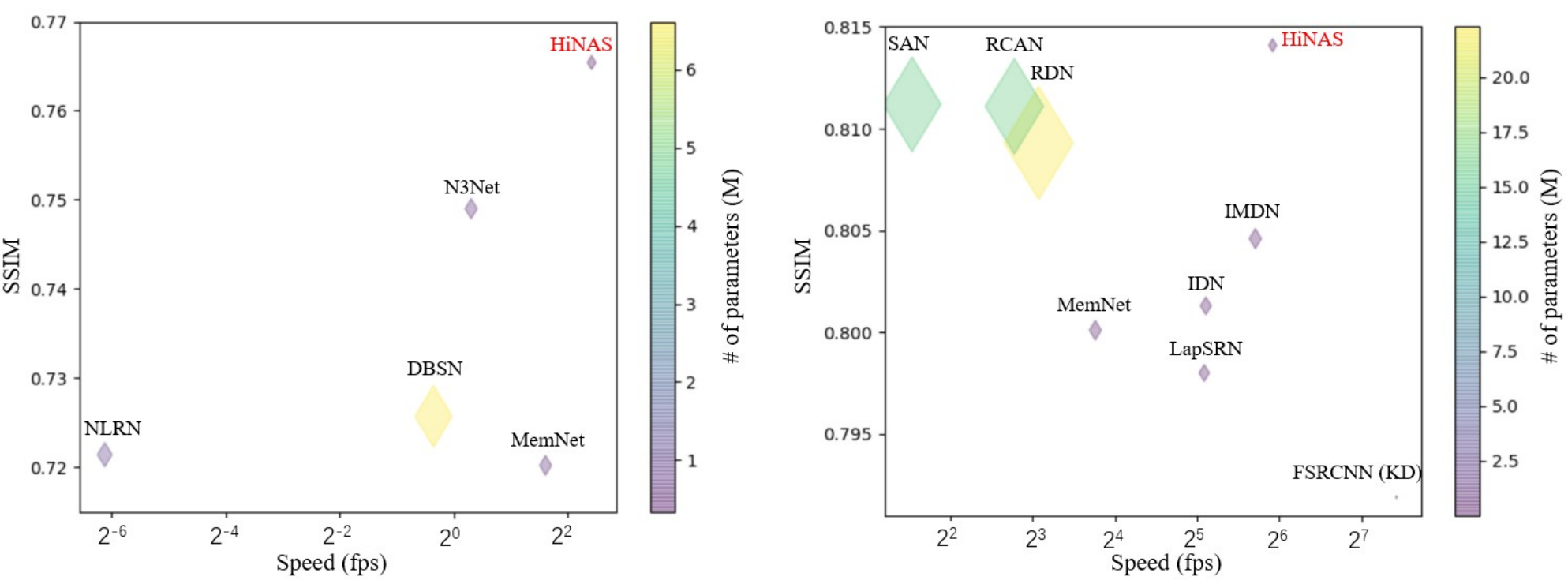}
\end{center}
\caption{Trade-off between inference speed and performance. (a) denoising results of test on BSD200 when $\sigma=50$. (b) 
$\times$3 super-resolution results of test on BSD100. Marker size and color encode the number of parameters.
HiNAS shows superior performance compared with a few recent methods.
}
\label{fig:scatter_map}
\end{figure*}

Now we compare the \OurMethod designed networks with a number of recent methods and use PSNR and SSIM to quantitatively measure the restoration performance of those methods. Table~\ref{Table: BSD500} shows the comparison results of denoising on BSD500 and Figure~\ref{fig:comparison_sota_dn}. 

Table~\ref{Table: BSD500} shows that N3Net and \OurMethod beat other models by a clear margin. Our proposed \OurMethod achieves the best performance when $\sigma$ is set to 50 and 70. When the noise level $\sigma$ is set to 30, the SSIM of NLRN is slightly higher (0.002) than that of our \OurMethod, but the PSNR of NLRN is much lower (nearly 1dB) than that of \OurMethod. 

\textit{Overall, our \OurMethod achieves better performance than others. In addition, compared with the second best model N3Net, the network designed by \OurMethod has fewer parameters and is faster in inference.} As listed in Table~\ref{Table: BSD500}, the \OurMethod designed network has 0.34M parameters, which is 47.89\% that of N3Net and 30.91\% that of E-CAE. Compared with N3Net, the \OurMethod designed network reduces the inference time on the test set of BSD500 by \textbf{79.92\%}. 

The comparison results of $\times$ 4 super-resolution experiments  are listed in Table~\ref{Table: comparison_with_NASs_sr} and more super-resolution experimental results of $\times$2, $\times$3 and $\times$4 are listed in Table \ref{tab: SR_experiments}. Compared with parameter heavy models which have more than 10 M parameters, \OurMethod achieves lower PSNR and highly competitive SSIM, while having much faster inference speed and much fewer parameters. For instance, the network designed by \OurMethod  has only  0.33 M parameters, which is about  \nicefrac{1}{47} that of SAN. The inference speed is 
{20.97$\times$} that of SAN. Compared with fast super-resolution models, \OurMethod achieves equal performance, while having faster inference speed. The recent method of 
FSRCNN (KD)
leverages distillation to improve the performance of FSRCNN. This method has fewer parameters and faster inference speed compared with our \OurMethod. However, the PSNR and SSIM of FSRCNN (KD) is much lower than ours and that of other fast super-resolution methods. So, we believe \OurMethod is still comparable compared with FSRCNN (KD) when taking both inference speed and performance into consideration. Two examples are presented in Figure~\ref{fig:comparison_sota_sr}, from which we can draw two conclusions: 1) all three CNN based super-resolution models achieve significantly better visual results than bicubic; 2) slightly higher PSNR can not guarantee plausible visual results. In the first row, both SAN and IMDN achieve higher PSNR compared to \OurMethod, but we can not see any difference from the results. In the second row, the PSNR and SSIM of the result of SAN are slightly higher than that of \OurMethod, while the visual results of SAN and \OurMethod are still very close, and it is hard to say which one is better. In summary, in terms of visual results, \OurMethod is highly competitive with other state-of-the-art methods. 

\subsection{Trade-off between Inference Speed and Accuracy} 

We visualize in Figure~\ref{fig:scatter_map} the performance comparison of our \OurMethod and the state of the art in terms of the inference speed, SSIM and the number of parameters. 

For image denoising, our \OurMethod achieves %
advantages compared with other denoising networks. For image super-resolution, our \OurMethod achieves the best trade-off between inference speed and performance. In terms of the inference speed, our \OurMethod is only slower than FSRCNN (KD), but the performance of \OurMethod are much better than that of FSRCNN (KD). Compared with other fast super-resolution networks, our \OurMethod achieve higher SSIM, while having much faster inference speed and fewer parameters.

\begin{table*}[h]
\caption{Super-resolution experiments.}
\label{tab: SR_experiments}
\footnotesize
\renewcommand\arraystretch{1.0}
\begin{center}
{
\begin{tabular}{l|c|c|c|cc|cc|cc|cc}
\hline
\multirow{2}*{Method}  & \multirow{2}*{Scale} & \multirow{2}*{Param.\ i (M)} & time & \multicolumn{2}{c|}{Set5} & \multicolumn{2}{c|}{Set14} & \multicolumn{2}{c|}{BSD100} & \multicolumn{2}{c}{Urban100} \\
   &   &   & costs (s) &  PSNR  & SSIM &  PSNR  & SSIM &  PSNR  & SSIM &  PSNR  & SSIM \\
\hline
\hline
EDSR      & $\times$2  & 43.00 &  & 38.11 & 0.9602 & 33.92 & 0.9195 & 32.32 & 0.9013 & 32.93 & 0.9351 \\
SRMD      & $\times$2 & 1.50 &  & 37.79 & 0.9601 & 33.32 & 0.9159 & 32.05 & 0.8985 & 31.33 & 0.9204  \\
DBPN      & $\times$2  & 10.05 &  & 38.09 & 0.9600 & 33.85 & 0.9190 & 32.27 & 0.9000 & 32.55 & 0.9324  \\
RDN       & $\times$2  & 22.30 &  & 38.24 & 0.9614 & 34.01 & 0.9212 & 32.34 & 0.9017 & 32.89 & 0.9353  \\
RCAN      & $\times$2  & 16.00 &  & 38.27 & 0.9614 & 34.11 & 0.9216 & 32.41 & 0.9026 & 33.34 & 0.9384  \\
SAN       & $\times$2  & 15.70 &  & 38.31 & 0.9620 & 34.07 & 0.9213 & 32.42 & 0.9028 & 33.10 & 0.9370  \\
RFANet    & $\times$2  & 11.00 &  & 38.26 & 0.9615 & 34.16 & 0.9220 & 32.41 & 0.9026 & 33.33 & 0.9026  \\  
\hline
bicubic   & $\times$2 & - & - & 33.66 & 0.9299 & 30.24 & 0.8688 & 29.56 & 0.8431 & 26.88 & 0.8403  \\
FSRCNN(KD)& $\times$2  & \textcolor{red}{0.013}  & \textcolor{red}{0.67} & 37.33 & 0.9576 & 32.79 & 0.9105 & 31.65 & 0.8926 & 30.24 & 0.9071  \\
VDSR      & $\times$2  & 0.67 &  & 37.53 & 0.9590 & 33.05 & 0.9130 & 31.90 & 0.8960 & 30.77 & 0.9140  \\
MemNet    & $\times$2  & 0.68 & 8.50 & 37.78 & 0.9597 & 33.28 & 0.9142 & 32.08 & 0.8978 & \textcolor{blue}{31.31} & 0.9195  \\
FALSR-B   & $\times$2  & 0.33 & - & 37.61 & 0.9582 & 33.29 & 0.9143 & 31.97 & 0.8967 & 31.28 & 0.9191  \\
IDN       & $\times$2  & 0.55 & 5.62 & \textcolor{blue}{37.83} & 0.9600 & \textcolor{blue}{33.30} & 0.9148 & \textcolor{blue}{32.08} & 0.8985 & 31.27 & 0.9196  \\
IMDN      & $\times$2  & 0.69 & 2.47 & \textcolor{red}{38.00} & \textcolor{blue}{0.9605} & \textcolor{red}{33.63} & \textcolor{blue}{0.9177} & \textcolor{red}{32.19} & \textcolor{blue}{0.8996} & \textcolor{red}{32.17} & \textcolor{red}{0.9283}  \\ 
\OurMethod & $\times$2 & \textcolor{blue}{0.23-0.28} & \textcolor{blue}{2.04} & 37.65 & \textcolor{red}{0.9608} & 32.95 & \textcolor{red}{0.9191} & 31.97 & \textcolor{red}{0.9036} & 31.29 & \textcolor{blue}{0.9236}  \\
\hline
\hline

SPSR      & $\times$4 &  24.80  & 16.47 & 30.40 & 0.8627 & 26.64 & 0.7930 & 25.51 & 0.6576 & 29.41 & 0.8537 \\
EDSR      & $\times$4 & 43.00 &   -  & 32.46 & 0.8968 & 28.80 & 0.7876 & 27.71 & 0.7420 & 26.64 & 0.8033  \\
SRMD      & $\times$4 & 1.50  & - & 31.96 & 0.8925 & 28.35 & 0.7787 & 27.49 & 0.7337 & 25.68 & 0.7731  \\
DBPN      & $\times$4 & 10.00   &  -  & 32.47 & 0.8980 & 28.82 & 0.7860 & 27.72 & 0.7400 & 26.38 & 0.7946  \\
RDN       & $\times$4 & 22.30 & 10.71 & 32.47 & 0.8990 & 28.81 & 0.7871 & 27.72 & 0.7419 & 26.61 & 0.8028  \\
RCAN      & $\times$4 & 16.00 &  13.12  & 32.62 & 0.9001 & 28.86 & 0.7888 & 27.76 & 0.7435 & 26.82 & 0.8087  \\
SAN       & $\times$4 & 15.70 & 31.03 & 32.64 & 0.9003 & 28.92 & 0.7888 & 27.78 & 0.7436 & 26.79 & 0.8068  \\
RFANet    & $\times$4 & 11.00 & -  & 32.66 & 0.9004 & 28.88 & 0.7894 & 27.79 & 0.7442 & 26.92 & 0.8112 \\
\hline
bicubic   & $\times$4 & - & - & 28.42 & 0.8104 & 26.00 & 0.7027 & 25.96 & 0.6675 & 23.14 & 0.6577  \\
FSRCNN (KD)& $\times$4 & \textcolor{red}{0.013}  & \textcolor{red}{0.54} & 30.95 & 0.8759 & 27.77 & 0.7615 & 27.08 & 0.7188 & 24.82 & 0.7393  \\
VDSR      & $\times$4 & 0.67 & - & 31.35 & 0.8830 & 28.02 & 0.7680 & 27.29 & 0.0726 & 25.18 & 0.7540  \\
LapSRN    & $\times$4 & 0.50 & 3.68 & 31.54 & 0.8852 & 28.09 & 0.7700 & 27.32 & 0.7275 & 25.21 & 0.7562 \\
MemNet    & $\times$4 & 0.68 &  6.86  & 31.74 & 0.8893 & \textcolor{blue}{28.26} & 0.7723 & 27.40 & 0.7281 & 25.50 & 0.7630  \\
IDN       & $\times$4 & 0.55  &  2.47 & \textcolor{blue}{31.82} & 0.8903 & 28.25 & 0.7730 & \textcolor{blue}{27.41} & 0.7297 & 25.41 & 0.7632  \\
IMDN      & $\times$4 & 0.72  &  1.72 & \textcolor{red}{32.21} & \textcolor{red}{0.8948} & \textcolor{red}{28.58} & \textcolor{blue}{0.7811} & \textcolor{red}{27.56} & \textcolor{blue}{0.7353} & \textcolor{red}{26.04} & \textcolor{red}{0.7838}  \\
\OurMethod & $\times$4 & \textcolor{blue}{0.29-0.33} & \textcolor{blue}{1.45} & 31.72 &	\textcolor{blue}{0.8926} & 27.85 & \textcolor{red}{0.7881}	& 27.35	& \textcolor{red}{0.7465} & \textcolor{blue}{25.51} & \textcolor{blue}{0.7779}  \\
\hline
\end{tabular}
}
\end{center}
\end{table*}

\section{Conclusion}

In this study, we have proposed \OurMethod, an attempt of using NAS to automatically design effective network architectures for two low-level image restoration tasks, image denoising, and super-resolution. \OurMethod builds a hierarchical search space including the inner search space and outer search space, which can search  for the inner cell topological structures and outer layer widths, respectively.

We have proposed the layer-wise architecture sharing strategy to improve the flexibility of the search space, improving performance. We have also proposed a cell-sharing strategy to save memory, accelerating the search process. Finally, instead of searching for networks to learning the restoration result directly, our \OurMethod design networks to learning the residual information between low-quality images and high-quality images.  \OurMethod is both memory and computation efficient, taking only less than $\nicefrac{1}{6}$ days to search using a single GPU. Extensive experimental results demonstrate that the proposed \OurMethod achieves highly competitive performance compared with state-of-the-art models while having fewer parameters and faster inference speed. We believe that the proposed method can be applied to many other low-level image processing tasks.

\bibliographystyle{spbasic}      %
\bibliography{ref}

\end{document}